\documentclass[runningheads]{llncs}

\usepackage{eccv}

\usepackage{eccvabbrv}

\usepackage{graphicx}
\usepackage{booktabs}
\usepackage{microtype}
\usepackage{graphicx}
\usepackage{subcaption}
\usepackage{multirow}
\usepackage{booktabs} 
\usepackage{tabularx,array}
\usepackage{arydshln}
\usepackage{romannum}

\usepackage[textsize=tiny]{todonotes}
\usepackage[commandnameprefix=always]{changes}

\usepackage[accsupp]{axessibility}  

\usepackage[pagebackref,breaklinks,colorlinks,citecolor=eccvblue]{hyperref}

\usepackage{orcidlink}

\begin{document}

\title{An Empirical Study of World Model Quantization} 

\titlerunning{An Empirical Study of World Model Quantization}


\author{Zhongqian Fu$^{*}$ \and Tianyi Zhao$^{*}$ \and Kai Han$^{\dagger}$ \and \\
Hang Zhou \and Xinghao Chen \and Yunhe Wang}

\authorrunning{Zhongqian Fu et al.}

\institute{Huawei Noah's Ark Lab \\
\email{\{fuzhongqian, zhaotianyi13, kai.han\}@huawei.com}}

\maketitle
\footnotetext[1]{Equal contribution}
\footnotetext[4]{Project lead}

\begin{abstract}
World models learn an internal representation of environment dynamics, enabling agents to simulate and reason about future states within a compact latent space for tasks such as planning, prediction, and inference. However, running world models rely on hevay computational cost and memory footprint, making model quantization essential for efficient deployment. To date, the effects of post-training quantization (PTQ) on world models remain largely unexamined. In this work, we present a systematic empirical study of world model quantization using DINO-WM as a representative case, evaluating diverse PTQ methods under both weight-only and joint weight-activation settings.
We conduct extensive experiments on different visual planning tasks across a wide range of bit-widths, quantization granularities, and planning horizons up to 50 iterations. Our results show that quantization effects in world models extend beyond standard accuracy and bit-width trade-offs: group-wise weight quantization can stabilize low-bit rollouts, activation quantization granularity yields inconsistent benefits, and quantization sensitivity is highly asymmetric between encoder and predictor modules. Moreover, aggressive low-bit quantization significantly degrades the alignment between the planning objective and task success, leading to failures that cannot be remedied by additional optimization.
These findings reveal distinct quantization-induced failure modes in world model-based planning and provide practical guidance for deploying quantized world models under strict computational constraints. The code will be available
at \url{https://github.com/huawei-noah/noah-research/tree/master/QuantWM}.

\end{abstract}

\section{Introduction}
In the field of artificial intelligence, a World Model is an internal, predictive representation of an environment that allows an AI system to understand and simulate how that environment works~\cite{lecun2022path,zhu2024sora}.
World models play an increasingly important role in embodied decision-making by enabling agents to predict and reason about future environment dynamics. 
These models are commonly employed for planning from visual observations through iterative latent rollouts, and have demonstrated strong performance across a wide range of control and manipulation tasks. 
In this work, we focus on world models for visual planning and adopt DINO-WM~\cite{zhou2024dino} as a representative instantiation. It is worth noting that our approach can also be transferred to other world model architectures.

Despite their strong performance, planning paradigms based on world models generally rely on repeated model inference across multiple planning iterations. 
At each planning step, the world model is invoked multiple times to evaluate candidate trajectories, causing both computational cost and memory footprint to scale with the planning horizon. 
This repeated inference poses a significant efficiency challenge in practical deployment scenarios, such as robotics~\cite{wu2023daydreamer} and large-scale simulation~\cite{brooks2024video}. Although reduced-precision inference such as FP16 is commonly adopted to alleviate computational and memory overhead, it remains insufficient for long-horizon planning where inference is executed tens of times per episode.

This limitation motivates the exploration of more aggressive model compression techniques. Among them, post-training quantization (PTQ) is particularly attractive due to its simplicity and the fact that it does not require retraining the model.
\begin{figure*}[!t]
    \centering
    \includegraphics[width=0.95\linewidth]{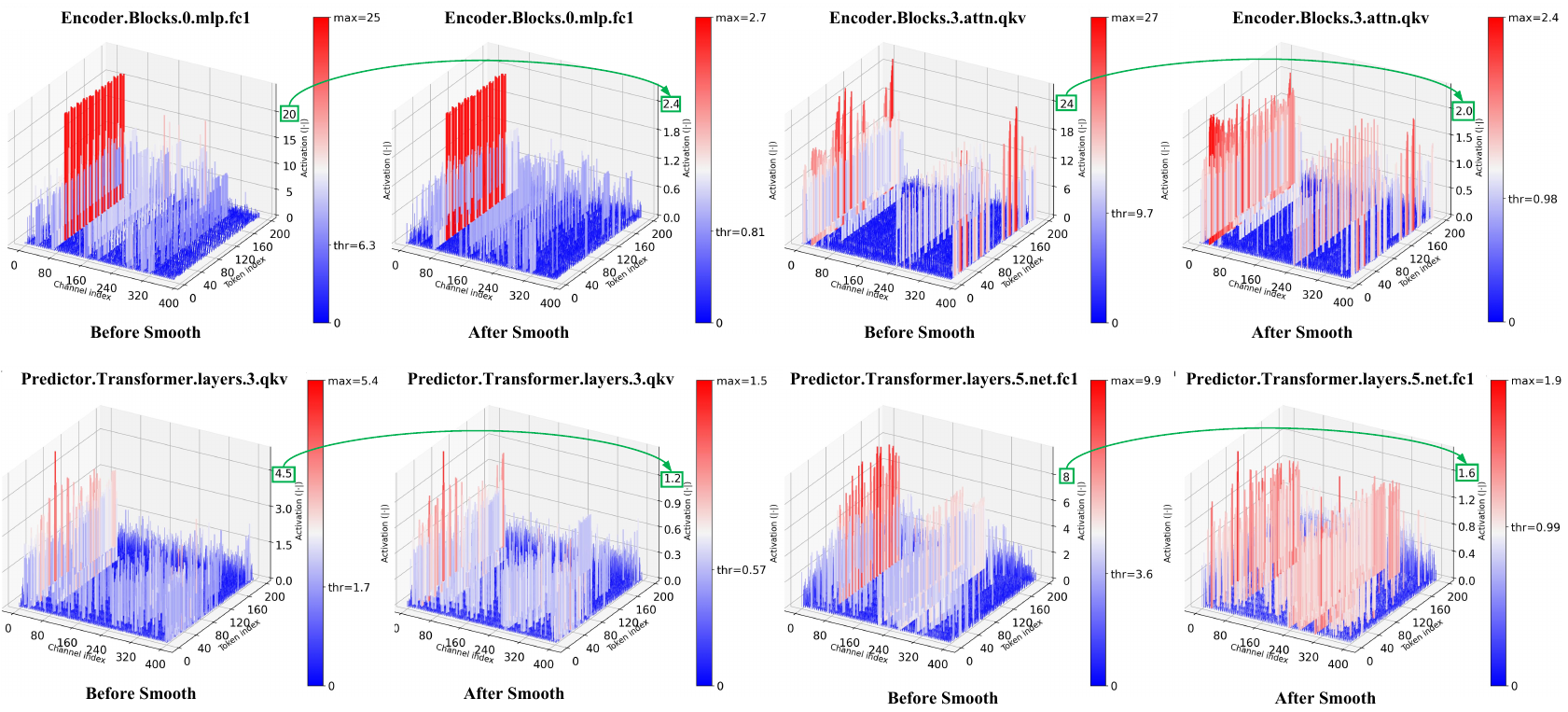}
    \caption{Activation outliers and scale imbalance in the encoder and predictor of DINO-WM before and after smoothing~\cite{xiao2023smoothquant}, illustrating numerical challenges for low-bit quantization in planning-based world models.}
    \label{fig:smooth}
    \vspace{-1em}
\end{figure*}
However, the behavior of PTQ has been predominantly studied for discriminative vision tasks~\cite{liu2021post} or autoregressive language modeling~\cite{xiao2023smoothquant}, but poorly understood in iterative planning on world models, where recurrent latent rollouts present unique challenges.
Unlike single-pass inference tasks, planning with world models involves temporal accumulation of errors, where small numerical perturbations introduced by quantization may compound across planning iterations and significantly affect long-horizon performance.
Moreover, different PTQ methods, ranging from simple rounding-based approaches to calibration-driven and activation-aware techniques, exhibit substantially different robustness characteristics, especially under low-bit configurations.
As a result, it remains unclear which quantization strategies are suitable for world model-based planning and how aggressively quantization can be applied without compromising planning reliability.

A key challenge in quantizing world models lies in their activation and representation characteristics.
As illustrated in Figure~\ref{fig:smooth}, the encoder and predictor in DINO-WM exhibit pronounced activation outliers and highly non-uniform scale distributions.
Such behavior amplifies the impact of quantization noise and makes numerical errors more likely to propagate across planning iterations.
While recent techniques such as activation smoothing can partially mitigate these effects, how quantization-induced distortions interact with long-horizon rollout dynamics remains unclear.

In this work, we conduct a systematic post-training quantization study on world models for planning to characterize the trade-offs between quantization efficiency and planning performance, using DINO-WM as a representative case study. We evaluate a diverse set of representative PTQ methods, including RTN~\cite{krishnamoorthi2018rtn}, OMSE~\cite{choukroun2019low}, AWQ~\cite{lin2024awq}, SmoothQuant~\cite{xiao2023smoothquant}, and OmniQuant~\cite{shao2023omniquant}, under both weight-only and joint weight-activation quantization settings. Our experiments cover a wide range of bit-widths and quantization granularities, and are evaluated on two embodied planning environments with planning horizons of up to 50 iterations. Specifically, this study seeks to answer the following questions:
\begin{itemize}
    \item How does aggressive low-bit quantization affect the performance and stability of world model in long-horizon planning, beyond standard accuracy and bit-width trade-offs?
    \item How do different PTQ strategies and granularities (e.g., per-channel vs. per-group weights, per-tensor vs. per-token activations) influence rollout dynamics and planning robustness?
    \item How sensitive are the encoder and predictor components of a world model to quantization noise, and which module constitutes the primary bottleneck for low-bit deployment?
\end{itemize}

Through extensive empirical evaluation, our work provides practical insights into quantizing world models. 
We hope these findings can serve as a useful reference for deploying DINO-WM and similar world models under strict computational constraints.

\section{Related Works}
\subsection{World Models}
World models aim to learn compact and predictive representations of environments to enable efficient planning and decision-making in reinforcement learning (RL) \cite{ha2018world, hafner2019planet, hafner2020dreamer, li2025surveywm}. These models typically combine latent state estimation, reward prediction, and policy learning within a unified framework. 
Early works like World Models \cite{ha2018world} introduced a combination of variational autoencoders (VAEs) and recurrent neural networks (RNNs) to model environment dynamics, while later approaches such as PlaNet \cite{hafner2019planet} and Dreamer \cite{hafner2020dreamer} replaced RNNs with deep latent variable models for improved scalability and performance. Recent advancements focus on integrating transformers \cite{vaswani2017attention} to capture long-term dependencies \cite{hafner2023dreamerv3} and incorporating multi-modal observations \cite{ge2024worldgpt}.

A notable recent development is DINO-WM \cite{zhou2024dino}, which leverages pre-trained visual features from self-supervised models (e.g., DINO \cite{caron2021dino, oquab2024dinov2}) to construct world models capable of zero-shot planning in unseen environments. 
By decoupling visual feature extraction from environment modeling, DINO-WM avoids the need for task-specific reward functions and instead relies on the semantic richness of pre-trained features to guide planning. 
This approach demonstrates strong generalization across diverse tasks and environments, particularly in scenarios where reward engineering is challenging. 

However, relying on pre-trained features introduces a fixed bottleneck, and the computational cost of maintaining high-resolution latent representations remains a major challenge for real-time deployment. Furthermore, due to the complexity of modeling long-term dynamics, scaling world models to high-fidelity visual environments still requires substantial computational resources\cite{li2025surveywm}. 

\subsection{Model Quantization}
Model quantization is a model compression technique that improves inference efficiency by converting full-precision parameters into low-precision integer representations, leading to direct computational accelaration and memory saving \cite{gupta2015deeplearninglimitednumerical, nagel2021whitepaper, gholami2021quantsurvey}. The quantization process is mathematically formalized as:
\begin{equation}
\setlength\abovedisplayskip{0.1cm}
\setlength\belowdisplayskip{0.1cm}
Q(\mathbf{x}) = \text{clip}(\lfloor\frac{\mathbf{x}}{s}\rceil + z, 0, 2^b - 1),
\label{eq: act quant}
\end{equation}
where $\mathbf{x}$ denotes the full-precision parameter, ${b}$ indicates the number of bits to be quantized, $\lfloor\cdot\rceil$ denotes the round-to-nearest operation, and the $\text{clip}$ function is used to restrict the rounded value to the specified range $(0, 2^b - 1)$.
Notably, ${s}$ is the scaling factor, and ${z}$ is the zero point, both determined by the boundary value of $\mathbf{x}$.

Among various quantization methods, PTQ has emerged as a dominant strategy for compressing deep learning models due to its compatibility with pre-trained systems and minimal reliance on retraining \cite{gholami2021quantsurvey}. PTQ leverages a small calibration dataset to optimize quantization parameters, minimizing the accuracy drop caused by low-precision inference. 
PTQ has been effectively applied to various neural networks, including CNNs \cite{krishnamoorthi2018rtn, choukroun2019low, li2021brecq, wei2023qdrop}, Vision Transformers \cite{liu2021post,yuan2024ptq4vit, liu2023noisyquant}, and Language Transformers \cite{xiao2023smoothquant, lin2024awq, shao2023omniquant, lin2024duquant, sun2025flatquant}.
Despite its demonstrated success, the application of quantization to world models remains largely unexplored, presenting a significant open challenge within the research community.
For instance, quantization errors in latent state transitions or reward prediction modules can accumulate over time, degrading planning performance. 
To bridge this gap, our research delves into the unique challenges of world modeling and provides a systematic empirical study of world model quantization.
We hope this research will inspire future studies on quantization strategies that explicitly consider long-horizon planning dynamics.


\begin{table*}[!t]
\centering
\caption{
3 to 8-bits weight-only PTQ results on the \texttt{Wall} dataset. \#W denotes the weight quantization bit-width, \#A denotes the activation quantization bit-width, and \#G denotes the group size. Noted that the AQG represents the Activation Quantization Granularity.}
\label{tab:1}
\resizebox{\linewidth}{!}{%

\begin{tabular}{cccccc ccccccc}
\toprule
\multirow{2}{*}{Dataset} 
& \multirow{2}{*}{Method} 
& \multirow{2}{*}{\#W} 
& \multirow{2}{*}{\#A} 
& \multirow{2}{*}{\#G} 
& \multirow{2}{*}{AQG}
& \multicolumn{7}{c}{Success Rate} \\
\cmidrule(lr){7-13}
& & & & & & 0 iters & 5 iters & 10 iters & 20 iters & 30 iters & 40 iters & 50 iters \\
\midrule
\multirow{25}{*}{\texttt{Wall}}

& FP32        & 32 & 32 & $\text{--}$ & $\text{--}$ & 0.48 & 0.94 & 0.94 & 0.94 & 0.94 & 0.94 & 0.94 \\
\cline{2-13}

& RTN         &  8 & 32 & $\text{--}$ & per-tensor & 0.52 & 0.92 & \textbf{0.94} & \textbf{0.94} & 0.94 & \textbf{0.96} & \textbf{0.96} \\
& OMSE        &  8 & 32 & $\text{--}$ & per-tensor & 0.54 & \textbf{0.94} & \textbf{0.94} & \textbf{0.94} & 0.94 & 0.94 & 0.94 \\
& AWQ         &  8 & 32 & $\text{--}$ & per-tensor & 0.44 & 0.90 & \textbf{0.94} & \textbf{0.94} & \textbf{0.96} & \textbf{0.96} & \textbf{0.96} \\
& OmniQuant   &  8 & 32 & $\text{--}$ & per-tensor & \textbf{0.56} & \textbf{0.94} & \textbf{0.94} & \textbf{0.94} & \textbf{0.96} & \textbf{0.96} & \textbf{0.96} \\
\cdashline{2-13}

& RTN         &  4 & 32 & $\text{--}$ & per-tensor & 0.00 & 0.18 & 0.22 & 0.24 & 0.30 & 0.40 & 0.42 \\
& OMSE        &  4 & 32 & $\text{--}$ & per-tensor & 0.04 & 0.58 & 0.68 & 0.76 & \textbf{0.80} & \textbf{0.82} & \textbf{0.82} \\
& AWQ         &  4 & 32 & $\text{--}$ & per-tensor & 0.04 & 0.50 & 0.62 & 0.66 & 0.70 & 0.72 & 0.72 \\
& OmniQuant   &  4 & 32 & $\text{--}$ & per-tensor & \textbf{0.10} & \textbf{0.70} & \textbf{0.74} & \textbf{0.78} & \textbf{0.80} & \textbf{0.82} & \textbf{0.82} \\
\cdashline{2-13}

& RTN         &  3 & 32 & $\text{--}$ & per-tensor & 0.00 & 0.00 & 0.00 & 0.00 & 0.00 & 0.00 & 0.00 \\
& OMSE        &  3 & 32 & $\text{--}$ & per-tensor & 0.00 & 0.00 & 0.00 & \textbf{0.06} & \textbf{0.10} & \textbf{0.14} & \textbf{0.14} \\
& AWQ         &  3 & 32 & $\text{--}$ & per-tensor & 0.00 & 0.00 & 0.00 & 0.00 & 0.02 & 0.02 & 0.02 \\
& OmniQuant   &  3 & 32 & $\text{--}$ & per-tensor & 0.00 & \textbf{0.02} & \textbf{0.02} & \textbf{0.06} & \textbf{0.10} & \textbf{0.14} & \textbf{0.14} \\
\cline{2-13}

& RTN         &  8 & 32 & 128 & per-tensor & \textbf{0.52} & 0.90 & 0.94 & 0.94 & 0.94 & 0.94 & 0.94 \\
& OMSE        &  8 & 32 & 128 & per-tensor & \textbf{0.52} & 0.90 & 0.94 & 0.94 & 0.94 & 0.94 & 0.94 \\
& AWQ         &  8 & 32 & 128 & per-tensor & 0.50 & \textbf{0.92} & \textbf{0.96} & \textbf{0.96} & \textbf{0.96} & \textbf{0.96} & \textbf{0.96} \\
& OmniQuant   &  8 & 32 & 128 & per-tensor & 0.46 & \textbf{0.92} & \textbf{0.96} & \textbf{0.96} & \textbf{0.96} & \textbf{0.96} & \textbf{0.96} \\
\cdashline{2-13}

& RTN         &  4 & 32 & 128 & per-tensor & 0.18 & 0.76 & 0.78 & 0.82 & 0.82 & 0.84 & 0.86 \\
& OMSE        &  4 & 32 & 128 & per-tensor & 0.08 & 0.54 & 0.54 & 0.60 & 0.62 & 0.62 & 0.62 \\
& AWQ         &  4 & 32 & 128 & per-tensor & 0.18 & 0.72 & 0.74 & 0.78 & 0.78 & 0.82 & 0.82 \\
& OmniQuant   &  4 & 32 & 128 & per-tensor & \textbf{0.20} & \textbf{0.92} & \textbf{0.94} & \textbf{0.94} & \textbf{0.94} & \textbf{0.94} & \textbf{0.94} \\
\cdashline{2-13}

& RTN         &  3 & 32 & 128 & per-tensor & 0.00 & 0.00 & 0.00 & 0.00 & 0.00 & 0.00 & 0.00 \\
& OMSE        &  3 & 32 & 128 & per-tensor & 0.00 & 0.00 & 0.00 & 0.00 & 0.00 & 0.00 & 0.00 \\
& AWQ         &  3 & 32 & 128 & per-tensor & 0.00 & 0.00 & 0.00 & 0.00 & 0.00 & 0.00 & 0.00 \\
& OmniQuant   &  3 & 32 & 128 & per-tensor & 0.00 & \textbf{0.02} & \textbf{0.02} & \textbf{0.02} & \textbf{0.02} & \textbf{0.04} & \textbf{0.04} \\

\bottomrule
\end{tabular}

}
\end{table*}

\section{Empirical Study}

\subsection{Experiment Settings}
\noindent\textbf{Base Model.} 
We conduct all experiments on DINO-WM~\cite{zhou2024dino}, a transformer-based world model designed for planning from visual observations. Following the original work, we use the publicly released model checkpoint and do not distinguish between different model scales. All quantization methods are applied to the same pretrained model to ensure fair comparison.

\noindent\textbf{Evaluation Environments.} 
We evaluate quantized world models on two representative visual planning environments: Wall and PushT, following the evaluation protocol of DINO-WM~\cite{zhou2024dino,fu2020d4rl,tassa2018deepmind}. 
These environments are drawn from standard embodied control benchmarks and exhibit different levels of dynamics complexity, covering both navigation and fine-grained manipulation tasks.
In both environments, the task is to reach a randomly sampled goal state specified by a target visual observation, starting from arbitrary initial states. 
Planning is performed by rolling out candidate action sequences using the world model and selecting actions that minimize the discrepancy between predicted future observations and the target observation.
Observations in all environments consist of RGB images. 
We refer readers to the DINO-WM paper for a full description of the environments and task specifications.

We report planning results over planning horizons ranging from 0 to 50 iterations, covering both short-horizon and long-horizon planning behaviors.

\noindent\textbf{Quantization Methods.} 
We evaluate several representative PTQ methods, covering both naive and calibration-based approaches that are widely adopted. Specifically, we consider the following methods:
\begin{itemize}
    \item \textbf{RTN}~\cite{krishnamoorthi2018rtn}: a uniform round-to-nearest quantization method that directly quantizes weights using affine quantization parameters.
    \item \textbf{OMSE}~\cite{choukroun2019low}: a calibration-based quantization method that determines quantization parameters by minimizing the mean squared error between floating-point and quantized layer outputs.
    \item \textbf{AWQ}~\cite{lin2024awq}: an activation-aware weight quantization method that identifies and preserves salient weight channels based on activation statistics, improving robustness under low-bit settings.
    \item \textbf{SmoothQuant}~\cite{xiao2023smoothquant}: a joint weight-activation quantization method that mitigates activation outliers by smoothing the scale distribution between weights and activations.
    \item \textbf{OmniQuant}~\cite{shao2023omniquant}: a calibration-driven quantization framework that jointly optimizes quantization parameters across layers, supporting both weight-only and weight-activation quantization.
\end{itemize}

\noindent\textbf{Quantization Configuration.}
We evaluate both weight-only and joint weight-activation quantization settings to comprehensively assess the impact of numerical precision on planning performance. For weight-only quantization, we consider 3-bit, 4-bit, and 8-bit weight precision. For joint weight-activation quantization, we evaluate the configurations including \texttt{W8A8}, \texttt{W6A6}, \texttt{W4A8}, and \texttt{W4A4}.

Quantization parameters are determined using a dedicated calibration dataset, which is strictly separated from the evaluation data. Calibration trajectories are collected by interacting with the environment using random seeds different from those used for evaluation, ensuring that no evaluation information is leaked during quantization. For calibration, we sample short planning trajectories with a fixed planning horizon of 2 iterations. The resulting observations and intermediate model activations are used to estimate quantization parameters for both weights and activations.

We further investigate the impact of quantization granularity by comparing per-channel and per-group quantization for weights, and per-tensor and per-token quantization for activations. For per-group quantization, we use a fixed group size of 128 across all experiments. Unless otherwise stated, symmetric quantization is used for weights, while asymmetric quantization is used for activations. All quantized models are evaluated using the same inference codebase to ensure that any observed performance differences are solely attributed to the quantization effects.

\begin{table*}[!t]
\centering
\caption{4 to 8-bits weight-only PTQ results on the \texttt{PushT} dataset.}
\label{tab:2}
\setlength{\tabcolsep}{0.2cm}
\resizebox{\linewidth}{!}{%

\begin{tabular}{cccccc ccccc}
\toprule
\multirow{2}{*}{Dataset} 
& \multirow{2}{*}{Method} 
& \multirow{2}{*}{\#W} 
& \multirow{2}{*}{\#A} 
& \multirow{2}{*}{\#G} 
& \multirow{2}{*}{AQG}
& \multicolumn{5}{c}{Success Rate} \\
\cmidrule(lr){7-11}
& & & & & & 0 iters & 5 iters & 10 iters & 20 iters & 30 iters \\
\midrule

\multirow{17}{*}{\texttt{PushT}}

& FP32        & 32 & 32 & $\text{--}$ & $\text{--}$ & 0.92 & 0.94 & 0.94 & 0.94 & 0.94 \\
\cline{2-11}

& RTN         &  8 & 32 & $\text{--}$ & per-tensor & 0.88 & \textbf{0.90} & \textbf{0.90} & \textbf{0.90} & \textbf{0.90} \\
& OMSE        &  8 & 32 & $\text{--}$ & per-tensor & \textbf{0.90} & \textbf{0.90} & \textbf{0.90} & \textbf{0.90} & \textbf{0.90} \\
& AWQ         &  8 & 32 & $\text{--}$ & per-tensor & 0.64 & 0.70 & 0.70 & 0.70 & 0.72 \\
& OmniQuant   &  8 & 32 & $\text{--}$ & per-tensor & 0.82 & 0.88 & 0.88 & 0.88 & 0.88 \\
\cdashline{2-11}

& RTN         &  4 & 32 & $\text{--}$ & per-tensor & \textbf{0.04} & \textbf{0.08} & \textbf{0.08} & \textbf{0.10} & \textbf{0.10} \\
& OMSE        &  4 & 32 & $\text{--}$ & per-tensor & 0.00 & 0.06 & \textbf{0.08} & 0.08 & 0.08 \\
& AWQ         &  4 & 32 & $\text{--}$ & per-tensor & 0.00 & 0.00 & 0.00 & 0.00 & 0.00 \\
& OmniQuant   &  4 & 32 & $\text{--}$ & per-tensor & 0.00 & 0.00 & 0.00 & 0.00 & 0.00 \\
\cline{2-11}

& RTN         &  8 & 32 & 128 & per-tensor & \textbf{0.88} & \textbf{0.90} & \textbf{0.90} & \textbf{0.90} & \textbf{0.90} \\
& OMSE        &  8 & 32 & 128 & per-tensor & 0.86 & 0.86 & 0.86 & 0.86 & 0.86 \\
& AWQ         &  8 & 32 & 128 & per-tensor & 0.58 & 0.72 & 0.72 & 0.72 & 0.72 \\
& OmniQuant   &  8 & 32 & 128 & per-tensor & 0.86 & 0.88 & 0.88 & 0.88 & 0.88 \\
\cdashline{2-11}

& RTN         &  4 & 32 & 128 & per-tensor & 0.16 & 0.38 & 0.40 & 0.44 & 0.44 \\
& OMSE        &  4 & 32 & 128 & per-tensor & 0.06 & 0.12 & 0.12 & 0.12 & 0.12 \\
& AWQ         &  4 & 32 & 128 & per-tensor & \textbf{0.32} & \textbf{0.50} & \textbf{0.50} & \textbf{0.50} & \textbf{0.50} \\
& OmniQuant   &  4 & 32 & 128 & per-tensor & 0.24 & 0.32 & 0.36 & 0.40 & 0.40 \\

\bottomrule
\end{tabular}

}
\end{table*}

\begin{figure*}[!t]
    \centering
    \includegraphics[width=0.9\linewidth]{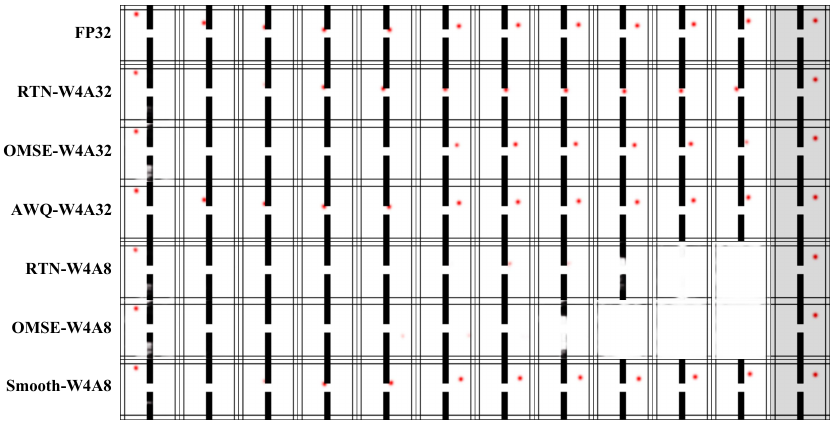}
    \includegraphics[width=0.9\linewidth]{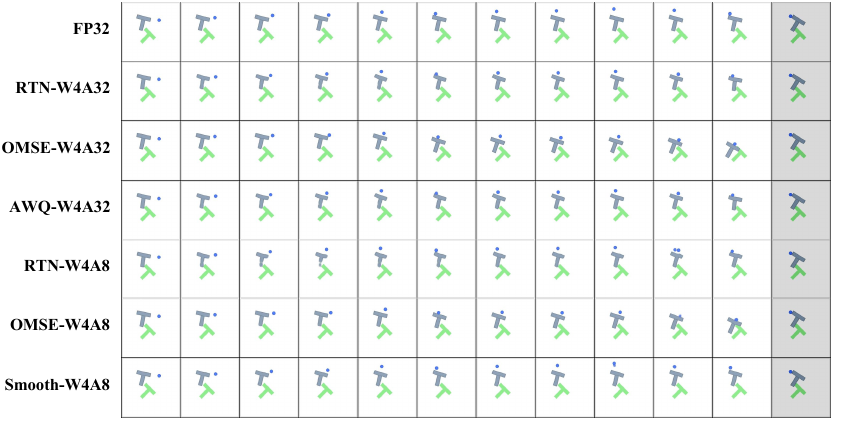}
    \caption{Open-loop rollouts of the world model under different quantization methods and bit-widths on WALL and Push-T. Given the first frame and a fixed action sequence, the model predicts future observations, which are reconstructed by the decoder. Results are shown for various quantization settings to illustrate their impact on long-horizon prediction quality. The top row shows the full-precision trajectories.}
    \label{fig:rollouts}
\end{figure*}

\begin{table*}[!t]
\centering
\caption{
4 to 8-bits weight-activation PTQ results on the \texttt{Wall} dataset. \#W denotes the weight quantization bit-width, \#A denotes the activation quantization bit-width, and \#G denotes the group size. Noted that the AQG represents the Activation Quantization Granularity.}
\label{tab:3}
\resizebox{\linewidth}{!}{%
\begin{tabular}{cccccc ccccccc}
\toprule
\multirow{2}{*}{Dataset} 
& \multirow{2}{*}{Method} 
& \multirow{2}{*}{\#W} 
& \multirow{2}{*}{\#A} 
& \multirow{2}{*}{\#G} 
& \multirow{2}{*}{AQG} 
& \multicolumn{7}{c}{Success Rate} \\
\cmidrule(lr){7-13}
& & & & & & 0 iters & 5 iters & 10 iters & 20 iters & 30 iters & 40 iters & 50 iters \\
\midrule

\multirow{33}{*}{\texttt{Wall}}
& FP32        & 32 & 32 & $\text{--}$ & $\text{--}$ & 0.48 & 0.94 & 0.94 & 0.94 & 0.94 & 0.94 & 0.94 \\
\cline{2-13}

& RTN         & 8 & 8 & $\text{--}$ & per-tensor & 0.48 & \textbf{0.94} & \textbf{0.96} & 0.96 & 0.96 & 0.96 & 0.96 \\
& OMSE        & 8 & 8 & $\text{--}$ & per-tensor & 0.44 & \textbf{0.94} & \textbf{0.96} & 0.96 & \textbf{0.98} & \textbf{0.98} & \textbf{0.98} \\
& SmoothQuant & 8 & 8 & $\text{--}$ & per-tensor & 0.46 & \textbf{0.94} & 0.94 & \textbf{0.98} & \textbf{0.98} & \textbf{0.98} & \textbf{0.98} \\
& OmniQuant   & 8 & 8 & $\text{--}$ & per-tensor & \textbf{0.50} & 0.92 & 0.94 & 0.96 & 0.96 & 0.96 & 0.96 \\
\cdashline{2-13}

& RTN         & 6 & 6 & $\text{--}$ & per-tensor & 0.24 & 0.78 & 0.80 & 0.84 & 0.86 & 0.86 & 0.88 \\
& OMSE        & 6 & 6 & $\text{--}$ & per-tensor & 0.26 & 0.84 & 0.88 & 0.88 & 0.90 & 0.90 & 0.90 \\
& SmoothQuant & 6 & 6 & $\text{--}$ & per-tensor & \textbf{0.32} & \textbf{0.90} & 0.90 & 0.90 & 0.90 & 0.90 & 0.90 \\
& OmniQuant   & 6 & 6 & $\text{--}$ & per-tensor & 0.20 & \textbf{0.90} & \textbf{0.92} & \textbf{0.92} & \textbf{0.94} & \textbf{0.94} & \textbf{0.94} \\
\cdashline{2-13}

& RTN         & 4 & 8 & $\text{--}$ & per-tensor & 0.00 & 0.12 & 0.20 & 0.28 & 0.30 & 0.36 & 0.38 \\
& OMSE        & 4 & 8 & $\text{--}$ & per-tensor & 0.00 & 0.30 & 0.42 & 0.60 & 0.64 & 0.76 & 0.84 \\
& SmoothQuant & 4 & 8 & $\text{--}$ & per-tensor & \textbf{0.18} & 0.68 & 0.72 & \textbf{0.82} & \textbf{0.84} & \textbf{0.84} & 0.84 \\
& OmniQuant   & 4 & 8 & $\text{--}$ & per-tensor & 0.14 & \textbf{0.70} & \textbf{0.76} & \textbf{0.82} & 0.82 & 0.82 & \textbf{0.88} \\
\cdashline{2-13}

& RTN         & 4 & 4 & $\text{--}$ & per-tensor & {0.00} & 0.00 & 0.04 & \textbf{0.08} & 0.08 & 0.08 & 0.08 \\
& OMSE        & 4 & 4 & $\text{--}$ & per-tensor & {0.00} & \textbf{0.06} & \textbf{0.06} & 0.06 & 0.06 & 0.06 & 0.06 \\
& SmoothQuant & 4 & 4 & $\text{--}$ & per-tensor & {0.00} & 0.02 & 0.02 & \textbf{0.08} & \textbf{0.10} & \textbf{0.12} & \textbf{0.12} \\
& OmniQuant   & 4 & 4 & $\text{--}$ & per-tensor & {0.00} & 0.00 & 0.00 & 0.02 & 0.04 & 0.04 & 0.06 \\
\cline{2-13}

& RTN         & 8 & 8 & $\text{--}$ & per-token & 0.44 & 0.92 & \textbf{0.96} & \textbf{0.96} & \textbf{0.96} & \textbf{0.96} & \textbf{0.96} \\
& OMSE        & 8 & 8 & $\text{--}$ & per-token & 0.48 & \textbf{0.96} & \textbf{0.96} & \textbf{0.96} & \textbf{0.96} & \textbf{0.96} & \textbf{0.96} \\
& SmoothQuant & 8 & 8 & $\text{--}$ & per-token & \textbf{0.52} & 0.92 & 0.94 & 0.94 & 0.94 & 0.94 & 0.94 \\
& OmniQuant   & 8 & 8 & $\text{--}$ & per-token & \textbf{0.52} & 0.94 & 0.94 & 0.94 & 0.94 & \textbf{0.96} & \textbf{0.96} \\
\cdashline{2-13}

& RTN         & 6 & 6 & $\text{--}$ & per-token & 0.30 & \textbf{0.82} & \textbf{0.88} & \textbf{0.92} & \textbf{0.92} & \textbf{0.92} & \textbf{0.92} \\
& OMSE        & 6 & 6 & $\text{--}$ & per-token & 0.28 & 0.78 & 0.84 & 0.84 & 0.84 & 0.84 & 0.84 \\
& SmoothQuant & 6 & 6 & $\text{--}$ & per-token & \textbf{0.32} & 0.80 & 0.82 & 0.88 & 0.88 & 0.88 & 0.88 \\
& OmniQuant   & 6 & 6 & $\text{--}$ & per-token & 0.26 & 0.76 & 0.78 & 0.80 & 0.82 & 0.84 & 0.86 \\
\cdashline{2-13}

& RTN         & 4 & 8 & $\text{--}$ & per-token & 0.00 & 0.08 & 0.10 & 0.20 & 0.28 & 0.32 & 0.34 \\
& OMSE        & 4 & 8 & $\text{--}$ & per-token & 0.00 & 0.48 & 0.60 & 0.70 & 0.74 & 0.74 & 0.76 \\
& SmoothQuant & 4 & 8 & $\text{--}$ & per-token & \textbf{0.14} & \textbf{0.74} & \textbf{0.78} & \textbf{0.82} & \textbf{0.84} & \textbf{0.86} & \textbf{0.88} \\
& OmniQuant   & 4 & 8 & $\text{--}$ & per-token & 0.10 & 0.68 & \textbf{0.78} & 0.80 & \textbf{0.84} & 0.84 & 0.86 \\
\cdashline{2-13}

& RTN         & 4 & 4 & $\text{--}$ & per-token & {0.00} & 0.00 & 0.00 & 0.04 & 0.06 & 0.06 & 0.06 \\
& OMSE        & 4 & 4 & $\text{--}$ & per-token & {0.00} & 0.00 & 0.00 & 0.00 & 0.00 & 0.00 & 0.00 \\
& SmoothQuant & 4 & 4 & $\text{--}$ & per-token & {0.00} & \textbf{0.02} & \textbf{0.04} & 0.04 & 0.04 & 0.04 & 0.04 \\
& OmniQuant   & 4 & 4 & $\text{--}$ & per-token & {0.00} & 0.00 & 0.02 & \textbf{0.06} & \textbf{0.08} & \textbf{0.08} & \textbf{0.08} \\

\bottomrule
\end{tabular}

}
\end{table*}

\begin{table*}[!t]
\centering
\caption{
4 to 8-bits weight-activation PTQ results on the \texttt{Wall} dataset (\#G=128).}
\label{tab:4}
\resizebox{\linewidth}{!}{%

\begin{tabular}{cccccc ccccccc}
\toprule
\multirow{2}{*}{Dataset}
& \multirow{2}{*}{Method}
& \multirow{2}{*}{\#W}
& \multirow{2}{*}{\#A}
& \multirow{2}{*}{\#G}
& \multirow{2}{*}{AQG}
& \multicolumn{7}{c}{Success Rate} \\
\cmidrule(lr){7-13}
& & & & & & 0 iters & 5 iters & 10 iters & 20 iters & 30 iters & 40 iters & 50 iters \\
\midrule

\multirow{29}{*}{\texttt{Wall}}
& FP32        & 32 & 32 & $\text{--}$ & $\text{--}$ & 0.48 & 0.94 & 0.94 & 0.94 & 0.94 & 0.94 & 0.94 \\
\cline{2-13}

& RTN         & 8 & 8 & 128 & per-tensor & 0.42 & \textbf{0.94} & 0.94 & \textbf{0.96} & \textbf{0.98} & \textbf{0.98} & \textbf{0.98} \\
& OMSE        & 8 & 8 & 128 & per-tensor & \textbf{0.60} & \textbf{0.94} & \textbf{0.96} & \textbf{0.96} & \textbf{0.98} & \textbf{0.98} & \textbf{0.98} \\
& SmoothQuant & 8 & 8 & 128 & per-tensor & 0.48 & 0.92 & 0.92 & 0.94 & 0.96 & 0.96 & 0.96 \\
& OmniQuant   & 8 & 8 & 128 & per-tensor & 0.42 & \textbf{0.94} & 0.94 & \textbf{0.96} & 0.96 & \textbf{0.98} & \textbf{0.98} \\
\cdashline{2-13}

& RTN         & 6 & 6 & 128 & per-tensor & 0.20 & 0.82 & 0.82 & 0.84 & 0.84 & 0.86 & 0.86 \\
& OMSE        & 6 & 6 & 128 & per-tensor & 0.26 & 0.80 & 0.82 & 0.84 & 0.84 & 0.86 & 0.86 \\
& SmoothQuant & 6 & 6 & 128 & per-tensor & \textbf{0.38} & \textbf{0.92} & \textbf{0.94} & 0.94 & 0.94 & 0.94 & 0.94 \\
& OmniQuant   & 6 & 6 & 128 & per-tensor & 0.26 & 0.88 & \textbf{0.94} & \textbf{0.96} & \textbf{0.96} & \textbf{0.96} & \textbf{0.96} \\
\cdashline{2-13}

& RTN         & 4 & 8 & 128 & per-tensor & \textbf{0.24} & 0.64 & 0.66 & 0.66 & 0.66 & 0.68 & 0.68 \\
& OMSE        & 4 & 8 & 128 & per-tensor & 0.10 & 0.50 & 0.54 & 0.60 & 0.64 & 0.64 & 0.64 \\
& SmoothQuant & 4 & 8 & 128 & per-tensor & 0.22 & \textbf{0.84} & \textbf{0.84} & \textbf{0.92} & \textbf{0.94} & \textbf{0.96} & \textbf{0.96} \\
&OmniQuant	&4&	8	&128 &per-tensor	&0.12 	&0.80 &	\textbf{0.84} 	&0.88 	&\textbf{0.94}& 	0.94 &	0.94 \\

\cdashline{2-13}

& RTN         & 4 & 4 & 128 & per-tensor & {0.00} & \textbf{0.02} & 0.06 & 0.08 & 0.12 & 0.12 & 0.12 \\
& OMSE        & 4 & 4 & 128 & per-tensor & {0.00} & \textbf{0.02} & 0.04 & 0.06 & 0.10 & 0.10 & 0.10 \\
& SmoothQuant & 4 & 4 & 128 & per-tensor & {0.00} & \textbf{0.02} & \textbf{0.10} & \textbf{0.14} & \textbf{0.14} & \textbf{0.14} & \textbf{0.14} \\
& OmniQuant   & 4 & 4 & 128 & per-tensor & {0.00} & \textbf{0.02} & 0.06 & 0.08 & 0.10 & 0.10 & 0.12 \\
\cline{2-13}

& RTN         & 8 & 8 & 128 & per-token & 0.42 & \textbf{0.94} & 0.94 & \textbf{0.96} & \textbf{0.98} & \textbf{0.98} & \textbf{0.98} \\
& OMSE        & 8 & 8 & 128 & per-token & \textbf{0.60} & \textbf{0.94} & \textbf{0.96} & \textbf{0.96} & \textbf{0.98} & \textbf{0.98} & \textbf{0.98} \\
& SmoothQuant & 8 & 8 & 128 & per-token & 0.48 & 0.92 & 0.94 & \textbf{0.96} & 0.96 & 0.96 & 0.96 \\
& OmniQuant   & 8 & 8 & 128 & per-token & 0.52 & \textbf{0.94} & 0.94 & 0.94 & 0.94 & 0.94 & 0.94 \\
\cdashline{2-13}

& RTN         & 6 & 6 & 128 & per-token & 0.20 & 0.82 & 0.82 & 0.84 & 0.84 & 0.84 & 0.84 \\
& OMSE        & 6 & 6 & 128 & per-token & 0.26 & 0.80 & 0.82 & 0.84 & 0.84 & 0.86 & 0.86 \\
& SmoothQuant & 6 & 6 & 128 & per-token & \textbf{0.38} & \textbf{0.92} & \textbf{0.94} & \textbf{0.94} & \textbf{0.94} & \textbf{0.94} & \textbf{0.94} \\
& OmniQuant   & 6 & 6 & 128 & per-token & 0.36 & 0.80 & 0.82 & 0.86 & 0.86 & 0.92 & 0.92 \\
\cdashline{2-13}

& RTN         & 4 & 8 & 128 & per-token & 0.24 & 0.64 & 0.66 & 0.66 & 0.66 & 0.66 & 0.68 \\
& OMSE        & 4 & 8 & 128 & per-token & 0.10 & 0.50 & 0.54 & 0.60 & 0.64 & 0.64 & 0.64 \\
& SmoothQuant & 4 & 8 & 128 & per-token & 0.22 & \textbf{0.84} & 0.84 & \textbf{0.92} & \textbf{0.94} & \textbf{0.94} & \textbf{0.96} \\
& OmniQuant   & 4 & 8 & 128 & per-token & \textbf{0.26} & \textbf{0.84} & \textbf{0.88} & 0.88 & 0.92 & 0.92 & 0.92 \\
\cdashline{2-13}

& RTN         & 4 & 4 & 128 & per-token & {0.00} & \textbf{0.02} & 0.06 & 0.08 & 0.12 & 0.12 & 0.12 \\
& OMSE        & 4 & 4 & 128 & per-token & {0.00} & \textbf{0.02} & 0.04 & 0.06 & 0.10 & 0.10 & 0.10 \\
& SmoothQuant & 4 & 4 & 128 & per-token & {0.00} & \textbf{0.02} & \textbf{0.10} & \textbf{0.14} & \textbf{0.14} & \textbf{0.14} & \textbf{0.14} \\
& OmniQuant   & 4 & 4 & 128 & per-token & {0.00} & 0.00 & 0.00 & 0.02 & 0.08 & 0.12 & 0.12 \\

\bottomrule
\end{tabular}

}
\end{table*}

\begin{table*}[!t]
\centering
\caption{
4 to 8-bits weight-activation PTQ results on the \texttt{PushT} dataset.}
\label{tab:5}
\setlength{\tabcolsep}{0.18cm}
\renewcommand{\arraystretch}{0.9}
\resizebox{\linewidth}{!}{%

\begin{tabular}{cccccc ccccc}
\toprule
\multirow{2}{*}{Dataset}
& \multirow{2}{*}{Method}
& \multirow{2}{*}{\#W}
& \multirow{2}{*}{\#A}
& \multirow{2}{*}{\#G}
& \multirow{2}{*}{AQG}
& \multicolumn{5}{c}{Success Rate} \\
\cmidrule(lr){7-11}
& & & & & & 0 iters & 5 iters & 10 iters & 20 iters & 30 iters \\
\midrule

\multirow{25}{*}{\texttt{PushT}}
& FP32        & 32 & 32 & $\text{--}$ & $\text{--}$ & 0.92 & 0.94 & 0.94 & 0.94 & 0.94 \\
\cline{2-11}

& RTN         & 8 & 8 & $\text{--}$ & per-tensor & 0.50 & 0.66 & 0.66 & 0.66 & 0.66 \\
& OMSE        & 8 & 8 & $\text{--}$ & per-tensor & 0.54 & 0.68 & 0.68 & 0.68 & 0.68 \\
& SmoothQuant & 8 & 8 & $\text{--}$ & per-tensor & \textbf{0.58} & \textbf{0.76} & \textbf{0.78} & \textbf{0.78} & \textbf{0.78} \\
& OmniQuant   & 8 & 8 & $\text{--}$ & per-tensor & \textbf{0.58} & 0.72 & 0.72 & 0.72 & 0.72 \\
\cdashline{2-11}

& RTN         & 6 & 6 & $\text{--}$ & per-tensor & 0.14 & 0.24 & 0.26 & 0.28 & 0.30 \\
& OMSE        & 6 & 6 & $\text{--}$ & per-tensor & \textbf{0.22} & \textbf{0.36} & \textbf{0.40} & \textbf{0.40} & \textbf{0.40} \\
& SmoothQuant & 6 & 6 & $\text{--}$ & per-tensor & 0.20 & 0.34 & 0.36 & 0.36 & 0.36 \\
& OmniQuant   & 6 & 6 & $\text{--}$ & per-tensor & 0.16 & 0.26 & 0.26 & 0.30 & 0.30 \\
\cdashline{2-11}

& RTN         & 4 & 8 & $\text{--}$ & per-tensor & \textbf{0.04} & \textbf{0.10} & \textbf{0.10} & \textbf{0.10} & \textbf{0.10} \\
& OMSE        & 4 & 8 & $\text{--}$ & per-tensor & 0.00 & 0.04 & 0.06 & 0.06 & 0.06 \\
& SmoothQuant & 4 & 8 & $\text{--}$ & per-tensor & 0.02 & 0.06 & 0.06 & 0.06 & 0.06 \\
& OmniQuant   & 4 & 8 & $\text{--}$ & per-tensor & 0.02 & 0.02 & 0.02 & 0.02 & 0.02 \\
\cline{2-11}

& RTN         & 8 & 8 & $\text{--}$ & per-token & 0.78 & \textbf{0.88} & \textbf{0.88} & \textbf{0.88} & \textbf{0.88} \\
& OMSE        & 8 & 8 & $\text{--}$ & per-token & 0.76 & 0.82 & 0.84 & 0.84 & 0.84 \\
& SmoothQuant & 8 & 8 & $\text{--}$ & per-token & \textbf{0.80} & 0.86 & \textbf{0.88} & \textbf{0.88} & \textbf{0.88} \\
& OmniQuant   & 8 & 8 & $\text{--}$ & per-token & 0.74 & 0.78 & 0.78 & 0.78 & 0.78 \\
\cdashline{2-11}

& RTN         & 6 & 6 & $\text{--}$ & per-token & 0.22 & 0.46 & 0.48 & 0.54 & 0.54 \\
& OMSE        & 6 & 6 & $\text{--}$ & per-token & \textbf{0.50} & \textbf{0.62} & \textbf{0.64} & \textbf{0.66} & \textbf{0.66} \\
& SmoothQuant & 6 & 6 & $\text{--}$ & per-token & 0.40 & 0.46 & 0.56 & 0.58 & 0.58 \\
& OmniQuant   & 6 & 6 & $\text{--}$ & per-token & 0.40 & 0.58 & 0.58 & 0.58 & 0.60 \\
\cdashline{2-11}

& RTN         & 4 & 8 & $\text{--}$ & per-token & \textbf{0.04} & \textbf{0.10} & \textbf{0.10} & \textbf{0.10} & \textbf{0.10} \\
& OMSE        & 4 & 8 & $\text{--}$ & per-token & 0.02 & \textbf{0.10} & \textbf{0.10} & \textbf{0.10} & \textbf{0.10} \\
& SmoothQuant & 4 & 8 & $\text{--}$ & per-token & 0.00 & 0.04 & 0.04 & 0.04 & 0.04 \\
& OmniQuant   & 4 & 8 & $\text{--}$ & per-token & 0.00 & 0.00 & 0.00 & 0.00 & 0.00 \\

\bottomrule
\end{tabular}

}
\end{table*}

\begin{table*}[!t]
\centering
\caption{
4 to 8-bits weight-activation PTQ results on the \texttt{PushT} dataset (\#G=128).}
\label{tab:6}
\setlength{\tabcolsep}{0.18cm}
\renewcommand{\arraystretch}{0.9}
\resizebox{\linewidth}{!}{%

\begin{tabular}{cccccc ccccc}
\toprule
\multirow{2}{*}{Dataset}
& \multirow{2}{*}{Method}
& \multirow{2}{*}{\#W}
& \multirow{2}{*}{\#A}
& \multirow{2}{*}{\#G}
& \multirow{2}{*}{AQG}
& \multicolumn{5}{c}{Success Rate} \\
\cmidrule(lr){7-11}
& & & & & & 0 iters & 5 iters & 10 iters & 20 iters & 30 iters \\
\midrule

\multirow{25}{*}{\texttt{PushT}}
& FP32        & 32 & 32 & $\text{--}$ & $\text{--}$ & 0.92 & 0.94 & 0.94 & 0.94 & 0.94 \\
\cline{2-11}

& RTN         & 8 & 8 & 128 & per-tensor & \textbf{0.68} & 0.72 & 0.72 & 0.72 & 0.74 \\
& OMSE        & 8 & 8 & 128 & per-tensor & 0.46 & 0.58 & 0.60 & 0.60 & 0.60 \\
& SmoothQuant & 8 & 8 & 128 & per-tensor & 0.62 & \textbf{0.78} & \textbf{0.80} & \textbf{0.80} & \textbf{0.80} \\
& OmniQuant   & 8 & 8 & 128 & per-tensor & 0.66 & 0.76 & 0.76 & 0.76 & 0.76 \\
\cdashline{2-11}

& RTN         & 6 & 6 & 128 & per-tensor & 0.14 & 0.36 & 0.38 & 0.38 & 0.38 \\
& OMSE        & 6 & 6 & 128 & per-tensor & 0.12 & 0.28 & 0.34 & 0.36 & 0.36 \\
& SmoothQuant & 6 & 6 & 128 & per-tensor & 0.34 & \textbf{0.50} & \textbf{0.52} & \textbf{0.54} & \textbf{0.54} \\
& OmniQuant   & 6 & 6 & 128 & per-tensor & \textbf{0.38} & 0.44 & 0.48 & 0.48 & 0.48 \\
\cdashline{2-11}

& RTN         & 4 & 8 & 128 & per-tensor & \textbf{0.14} & \textbf{0.24} & \textbf{0.28} & \textbf{0.28} & \textbf{0.34} \\
& OMSE        & 4 & 8 & 128 & per-tensor & 0.08 & 0.18 & 0.22 & 0.26 & 0.26 \\
& SmoothQuant & 4 & 8 & 128 & per-tensor & \textbf{0.14} & 0.18 & 0.20 & 0.20 & 0.20 \\
& OmniQuant   & 4 & 8 & 128 & per-tensor & 0.12 & 0.18 & 0.18 & 0.20 & 0.22 \\
\cline{2-11}

& RTN         & 8 & 8 & 128 & per-token & \textbf{0.78} & \textbf{0.86} & \textbf{0.86} & \textbf{0.86} & \textbf{0.86} \\
& OMSE        & 8 & 8 & 128 & per-token & 0.74 & 0.82 & 0.84 & 0.84 & 0.84 \\
& SmoothQuant & 8 & 8 & 128 & per-token & \textbf{0.78} & 0.84 & 0.84 & 0.84 & 0.84 \\
& OmniQuant   & 8 & 8 & 128 & per-token & 0.70 & 0.78 & 0.78 & 0.78 & 0.78 \\
\cdashline{2-11}

& RTN         & 6 & 6 & 128 & per-token & 0.32 & 0.46 & 0.48 & 0.48 & 0.48 \\
& OMSE        & 6 & 6 & 128 & per-token & \textbf{0.38} & 0.44 & 0.46 & 0.48 & 0.48 \\
& SmoothQuant & 6 & 6 & 128 & per-token & 0.26 & \textbf{0.52} & \textbf{0.52} & \textbf{0.52} & \textbf{0.54} \\
& OmniQuant   & 6 & 6 & 128 & per-token & 0.26 & 0.38 & 0.40 & 0.40 & 0.40 \\
\cdashline{2-11}

& RTN         & 4 & 8 & 128 & per-token & \textbf{0.32} & 0.28 & 0.30 & 0.30 & 0.30 \\
& OMSE        & 4 & 8 & 128 & per-token & 0.12 & \textbf{0.32} & 0.18 & 0.20 & 0.20 \\
& SmoothQuant & 4 & 8 & 128 & per-token & \textbf{0.32} & 0.30 & \textbf{0.34} & \textbf{0.34} & \textbf{0.36} \\
& OmniQuant   & 4 & 8 & 128 & per-token & 0.18 & 0.28 & 0.28 & 0.30 & 0.30 \\

\bottomrule
\end{tabular}

}
\end{table*}

\begin{table*}[!t]
\centering
\caption{Encoder and Predictor quantization bit-width ablation on the \texttt{Wall} dataset.}
\label{tab:7}
\resizebox{\linewidth}{!}{%

\begin{tabular}{cccccc ccccccc}
\toprule
\multirow{2}{*}{Dataset}
& \multirow{2}{*}{Method}
& \multirow{2}{*}{\#W\_Enc}
& \multirow{2}{*}{\#A\_Enc}
& \multirow{2}{*}{\#W\_Pred}
& \multirow{2}{*}{\#A\_Pred}
& \multicolumn{7}{c}{Success Rate} \\
\cmidrule(lr){7-13}
& & & & & & 0 iters & 5 iters & 10 iters & 20 iters & 30 iters & 40 iters & 50 iters \\
\midrule

\multirow{23}{*}{\texttt{Wall}}
& FP32 & 32 & 32 & 32 & 32& 0.48 & 0.94 & 0.94 & 0.94 & 0.94 & 0.94 & 0.94 \\ \cline{2-13}

& RTN & 8 & 8 & 8 & 8 & 0.48 & 0.94 & 0.96 & 0.96 & 0.96 & 0.96 & 0.96 \\
& RTN & 6 & 8 & 8 & 8 & 0.42 & 0.94 & 0.94 & 0.94 & 0.94 & 0.94 & 0.94 \\
& RTN & 6 & 6 & 8 & 8 & 0.18 & 0.74 & 0.78 & 0.80 & 0.80 & 0.80 & 0.80 \\
& RTN & 4 & 8 & 8 & 8 & 0.02 & 0.14 & 0.38 & 0.50 & 0.54 & 0.56 & 0.64 \\
& RTN & 4 & 6 & 8 & 8 & 0.00 & 0.08 & 0.22 & 0.38 & 0.44 & 0.54 & 0.56 \\
& RTN & 4 & 4 & 8 & 8 & 0.00 & 0.00 & 0.00 & 0.02 & 0.02 & 0.02 & 0.02 \\ \cdashline{2-13}

& RTN & 8 & 8 & 6 & 8 & 0.48 & 0.96 & 0.96 & 0.96 & 0.98 & 0.98 & 0.98 \\
& RTN & 8 & 8 & 6 & 6 & 0.44 & 0.94 & 0.96 & 0.96 & 0.96 & 0.96 & 0.96 \\
& RTN & 8 & 8 & 4 & 8 & 0.34 & 0.90 & 0.92 & 0.94 & 0.94 & 0.94 & 0.94 \\
& RTN & 8 & 8 & 4 & 6 & 0.26 & 0.90 & 0.94 & 0.96 & 0.96 & 0.96 & 0.96 \\
& RTN & 8 & 8 & 4 & 4 & 0.08 & 0.56 & 0.58 & 0.58 & 0.60 & 0.60 & 0.60 \\ \cline{2-13}

& SmoothQuant & 8 & 8 & 8 & 8 & 0.46 & 0.94 & 0.94 & 0.98 & 0.98 & 0.98 & 0.98 \\
& SmoothQuant & 6 & 8 & 8 & 8 & 0.48 & 0.94 & 0.94 & 0.94 & 0.94 & 0.94 & 0.94 \\
& SmoothQuant & 6 & 6 & 8 & 8 & 0.26 & 0.76 & 0.80 & 0.84 & 0.84 & 0.86 & 0.88 \\
& SmoothQuant & 4 & 8 & 8 & 8 & 0.18 & 0.74 & 0.78 & 0.78 & 0.80 & 0.80 & 0.82 \\
& SmoothQuant & 4 & 6 & 8 & 8 & 0.14 & 0.64 & 0.76 & 0.78 & 0.80 & 0.80 & 0.80 \\
& SmoothQuant & 4 & 4 & 8 & 8 & 0.00 & 0.00 & 0.00 & 0.00 & 0.00 & 0.00 & 0.00 \\ \cdashline{2-13}

& SmoothQuant & 8 & 8 & 6 & 8 & 0.52 & 0.92 & 0.94 & 0.94 & 0.94 & 0.94 & 0.96 \\
& SmoothQuant & 8 & 8 & 6 & 6 & 0.46 & 0.94 & 0.96 & 0.96 & 0.96 & 0.96 & 0.96 \\
& SmoothQuant & 8 & 8 & 4 & 8 & 0.40 & 0.94 & 0.94 & 0.94 & 0.94 & 0.94 & 0.96 \\
& SmoothQuant & 8 & 8 & 4 & 6 & 0.42 & 0.92 & 0.94 & 0.94 & 0.94 & 0.94 & 0.94 \\
& SmoothQuant & 8 & 8 & 4 & 4 & 0.40 & 0.70 & 0.70 & 0.72 & 0.74 & 0.74 & 0.74 \\

\bottomrule
\end{tabular}
}
\end{table*}

\begin{table*}[!t]
\centering
\caption{Encoder and Predictor quantization bit-width ablation on the \texttt{PushT} dataset.}
\label{tab:8}
\renewcommand{\arraystretch}{1.1}
\resizebox{\linewidth}{!}{%
\begin{tabular}{cccccc ccccc}
\toprule
\multirow{2}{*}{Dataset}
& \multirow{2}{*}{Method}
& \multirow{2}{*}{\#W\_Enc}
& \multirow{2}{*}{\#A\_Enc}
& \multirow{2}{*}{\#W\_Pred}
& \multirow{2}{*}{\#A\_Pred}
& \multicolumn{5}{c}{Success Rate} \\ 
\cmidrule(lr){7-11}
& & & & & & 0 iters & 5 iters & 10 iters & 20 iters & 30 iters \\
\midrule

\multirow{23}{*}{\texttt{PushT}}
& FP32 & 32 & 32 & 32 & 32 
& 0.92 & 0.94 & 0.94 & 0.94 & 0.94 \\ \cline{2-11}

& RTN & 8 & 8 & 8 & 8 & 0.50 & 0.66 & 0.66 & 0.66 & 0.66 \\
& RTN & 6 & 8 & 8 & 8 & 0.38 & 0.54 & 0.56 & 0.56 & 0.56 \\
& RTN & 6 & 6 & 8 & 8 & 0.04 & 0.12 & 0.12 & 0.14 & 0.14 \\
& RTN & 4 & 8 & 8 & 8 & 0.02 & 0.02 & 0.02 & 0.02 & 0.02 \\
& RTN & 4 & 6 & 8 & 8 & 0.00 & 0.02 & 0.02 & 0.02 & 0.04 \\
& RTN & 4 & 4 & 8 & 8 & 0.02 & 0.02 & 0.02 & 0.02 & 0.02 \\ \cdashline{2-11}

& RTN & 8 & 8 & 6 & 8 & 0.46 & 0.70 & 0.72 & 0.72 & 0.72 \\
& RTN & 8 & 8 & 6 & 6 & 0.46 & 0.66 & 0.66 & 0.66 & 0.66 \\
& RTN & 8 & 8 & 4 & 8 & 0.40 & 0.44 & 0.44 & 0.44 & 0.44 \\
& RTN & 8 & 8 & 4 & 6 & 0.48 & 0.50 & 0.54 & 0.54 & 0.54 \\
& RTN & 8 & 8 & 4 & 4 & 0.00 & 0.00 & 0.02 & 0.02 & 0.02 \\ \cline{2-11}

& SmoothQuant & 8 & 8 & 8 & 8 & 0.58 & 0.76 & 0.78 & 0.78 & 0.78 \\
& SmoothQuant & 6 & 8 & 8 & 8 & 0.60 & 0.68 & 0.70 & 0.74 & 0.74 \\
& SmoothQuant & 6 & 6 & 8 & 8 & 0.14 & 0.22 & 0.24 & 0.24 & 0.24 \\
& SmoothQuant & 4 & 8 & 8 & 8 & 0.00 & 0.00 & 0.00 & 0.00 & 0.00 \\
& SmoothQuant & 4 & 6 & 8 & 8 & 0.00 & 0.00 & 0.00 & 0.00 & 0.00 \\
& SmoothQuant & 4 & 4 & 8 & 8 & 0.00 & 0.00 & 0.00 & 0.00 & 0.00 \\ \cdashline{2-11}

& SmoothQuant & 8 & 8 & 6 & 8 & 0.58 & 0.76 & 0.76 & 0.78 & 0.78 \\
& SmoothQuant & 8 & 8 & 6 & 6 & 0.54 & 0.74 & 0.76 & 0.78 & 0.78 \\
& SmoothQuant & 8 & 8 & 4 & 8 & 0.40 & 0.70 & 0.70 & 0.70 & 0.70 \\
& SmoothQuant & 8 & 8 & 4 & 6 & 0.34 & 0.60 & 0.64 & 0.66 & 0.66 \\
& SmoothQuant & 8 & 8 & 4 & 4 & 0.00 & 0.00 & 0.00 & 0.00 & 0.00 \\

\bottomrule
\end{tabular}
}
\end{table*}

\subsection{Post-Training Quantization Experiments}
\label{sec:ptq_experiments}

We present a detailed analysis of PTQ effects on world model planning performance. 
In addition to success rate, we also investigated how quantization precision, quantization granularity, and different modules interact with long-horizon rollout dynamics on the \texttt{Wall} and \texttt{PushT} environments.

\paragraph{Per-Channel vs.\ Per-Group Weight Quantization.}
Tables~\ref{tab:1} and~\ref{tab:2} report weight-only quantization results under per-channel and per-group settings. At 8-bit precision, all evaluated methods achieve performance comparable to FP32 across planning horizons, indicating that moderate weight quantization introduces negligible distortion to the learned dynamics. 
When weight precision is reduced to 4-bit, performance degradation is already evident at short planning horizons, particularly at 0-5 iterations, under both per-channel and per-group quantization. On \texttt{Wall}, increasing the number of planning iterations leads to substantial recovery primarily under per-group quantization, indicating that the planner can partially compensate for quantization-induced transition bias when grouping is applied. For example, under 4-bit weight quantization with OmniQuant and per-group grouping (\#G = 128), the success rate improves from 0.20 at 0 iterations to 0.94 at 50 iterations, closely matching the FP32 baseline. \texttt{PushT} exhibits limited recoverability under per-channel 4-bit quantization, while per-group quantization (\#G = 128) enables only partial recovery.
However, under extreme precision reduction (3-bit), performance collapses across both datasets regardless of the quantization method or grouping strategy. In this regime, success rates remain near zero even as planning iterations increase, indicating that quantization noise overwhelms the learned transition dynamics. As a result, the stabilizing effect of group-wise quantization largely disappears, suggesting that grouping is only effective when the underlying latent structure is sufficiently preserved under quantization.

\paragraph{Per-Tensor vs.\ Per-Token Activation Quantization.}
Tables~\ref{tab:3},~\ref{tab:4},~\ref{tab:5}, and~\ref{tab:6} investigate the impact of activation quantization granularity under both per-channel and per-group weight quantization. Across both environments, per-token activation quantization does not consistently outperform per-tensor quantization, despite its more fine-grained scaling.
At moderate bit-widths (e.g., W8A8 and W6A6), per-token activation quantization occasionally matches or slightly exceeds per-tensor performance, particularly on \texttt{Wall}. However, these gains become inconsistent under lower precision and longer planning horizons. In several low-bit settings, per-token quantization introduces additional variability, leading to reduced success rates despite comparable short-horizon performance.
These results suggest that, in iterative planning settings, maintaining globally consistent activation scaling is often more critical than maximizing instantaneous representational flexibility. Fine-grained activation scaling alone is insufficient to guarantee stable long-horizon rollouts when quantization noise accumulates over time.

\paragraph{Quantization Sensitivity of Encoder and Predictor.}
Tables~\ref{tab:7} and~\ref{tab:8} isolate the effects of quantization on the encoder and predictor modules. A clear asymmetry emerges across both environments. Quantizing the encoder leads to rapid performance degradation, especially under low-bit configurations. Errors introduced at the representation level propagate through all subsequent rollout steps, resulting in persistent performance loss that cannot be corrected by additional planning iterations.
In contrast, the predictor exhibits greater robustness to quantization. Moderate predictor quantization primarily affects temporal consistency rather than representation quality, and its impact can often be partially mitigated by increasing the planning horizon. This behavior is particularly evident on \texttt{Wall}, where success rates recover as planning iterations increase when only the predictor is quantized.
These observations indicate that representation quality constitutes the dominant bottleneck in low-bit world model deployment, while moderate transition noise introduced by predictor quantization is comparatively more tolerable in planning-based evaluation.

\paragraph{Qualitative Rollouts and Planning Loss Behavior.}
Figure~\ref{fig:rollouts} provides qualitative evidence supporting the quantitative trends. On \texttt{Wall}, encoder quantization leads to visible degradation in reconstructed observations starting from the initial frame, consistent with representation-level distortion. On \texttt{PushT}, reconstructed observations often remain visually plausible even when success rates drop sharply, indicating that task failures are not primarily caused by immediate visual representation collapse.

\begin{figure*}[!t]
    \centering
    \begin{minipage}[b]{0.19\textwidth}
        \includegraphics[width=\textwidth]{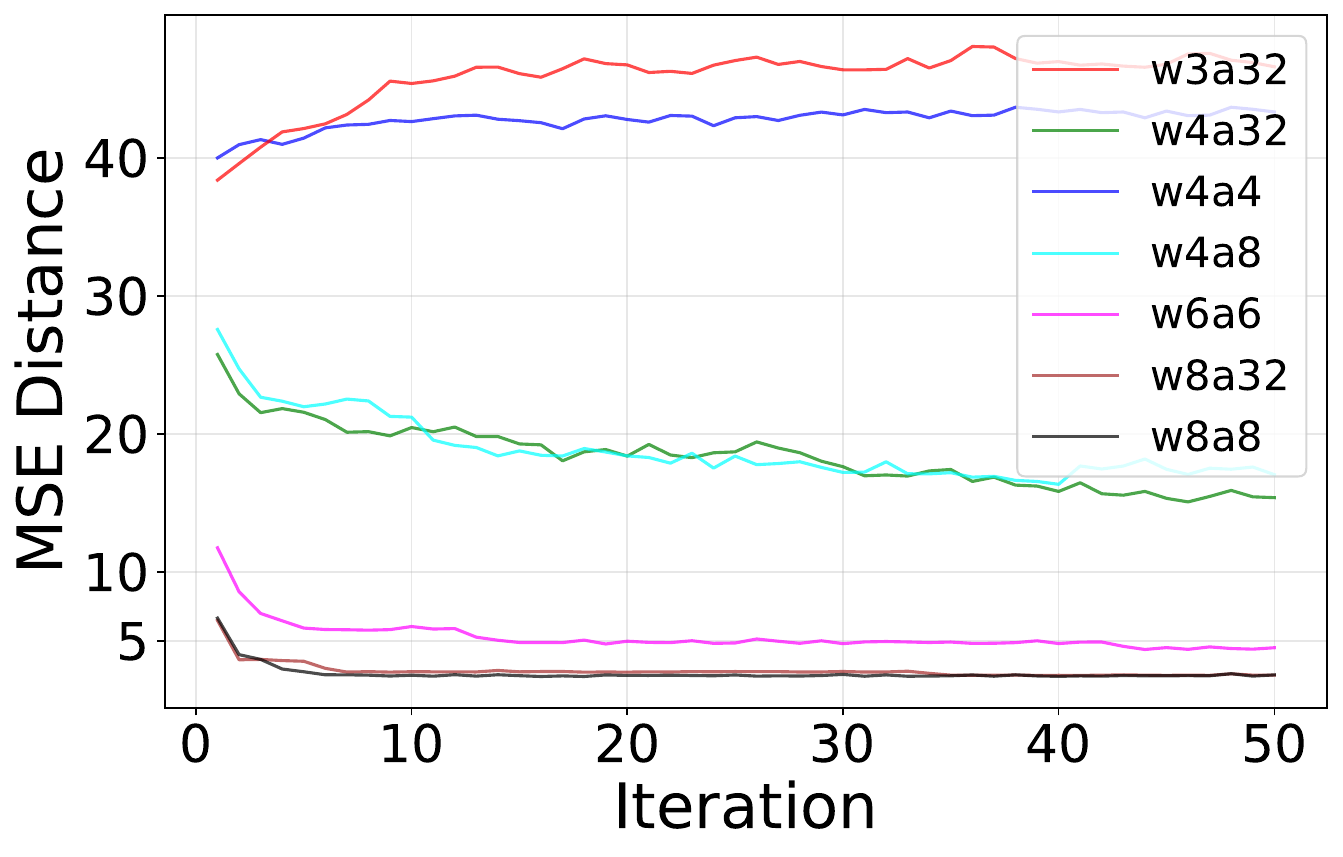}
        \subcaption{RTN}
    \end{minipage}
    \hspace{-0.01\textwidth}
    \begin{minipage}[b]{0.19\textwidth}
        \includegraphics[width=\textwidth]{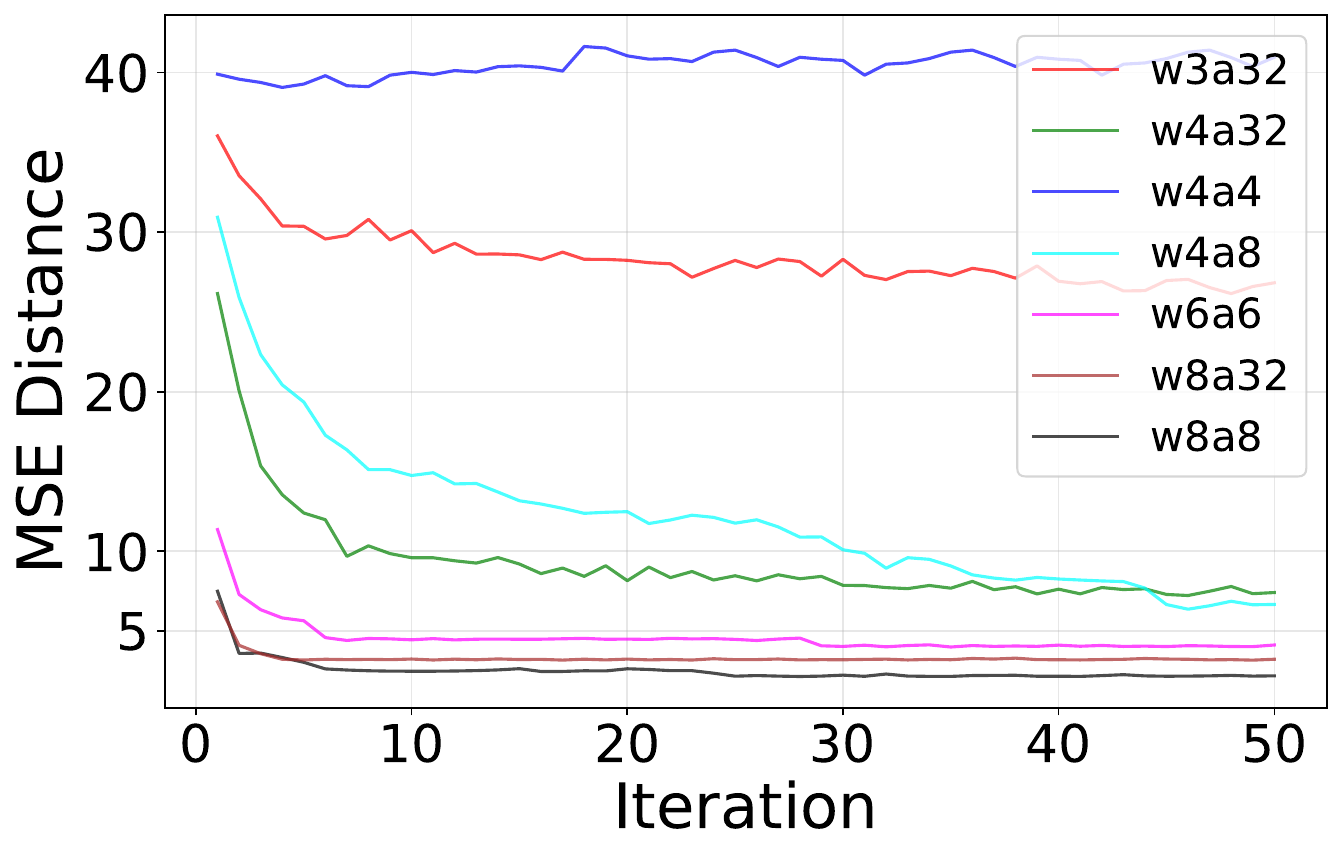}
        \subcaption{OMSE}
    \end{minipage}
    \hspace{-0.01\textwidth}
    \begin{minipage}[b]{0.19\textwidth}
        \includegraphics[width=\textwidth]{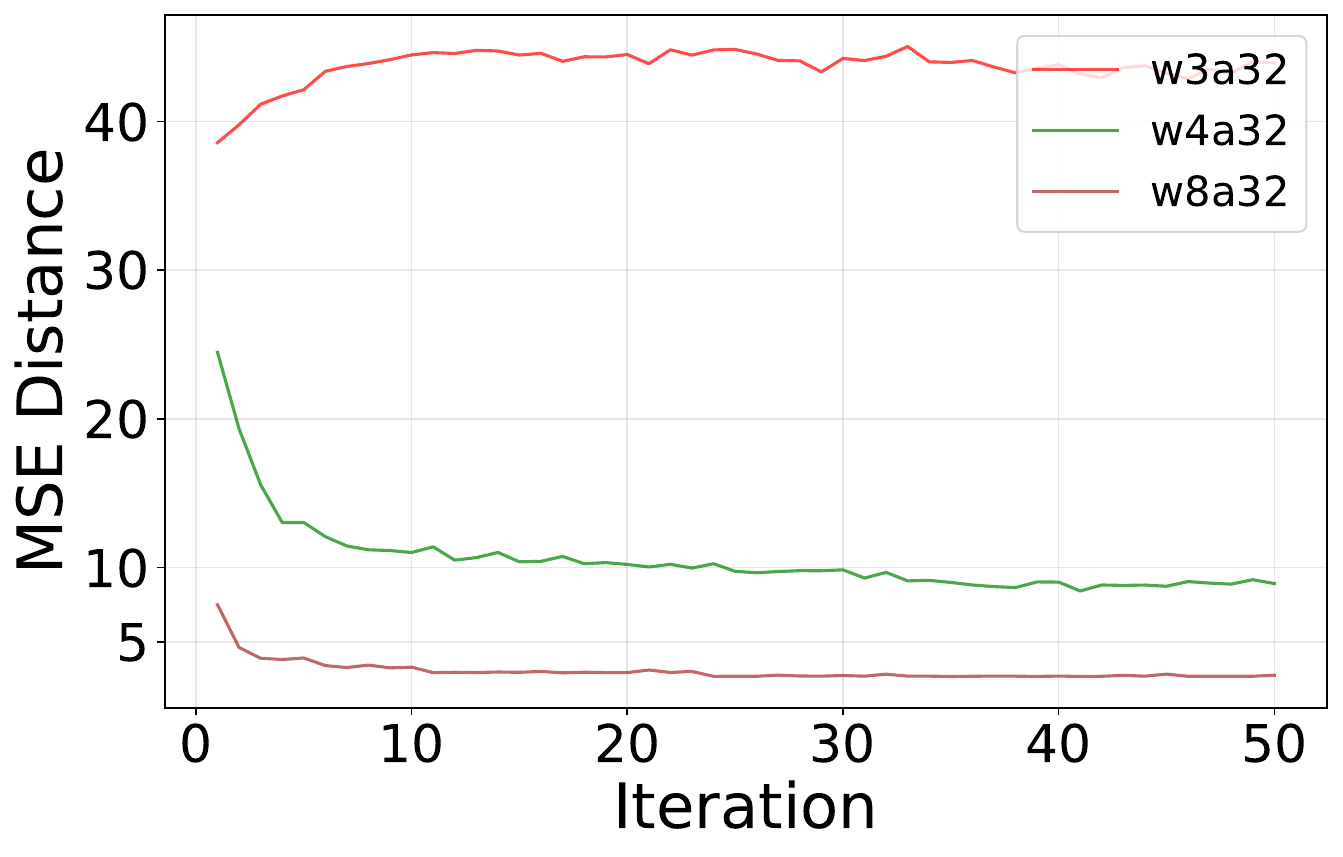}
        \subcaption{AWQ}
    \end{minipage}
    \hspace{-0.01\textwidth}
    \begin{minipage}[b]{0.19\textwidth}
        \includegraphics[width=\textwidth]{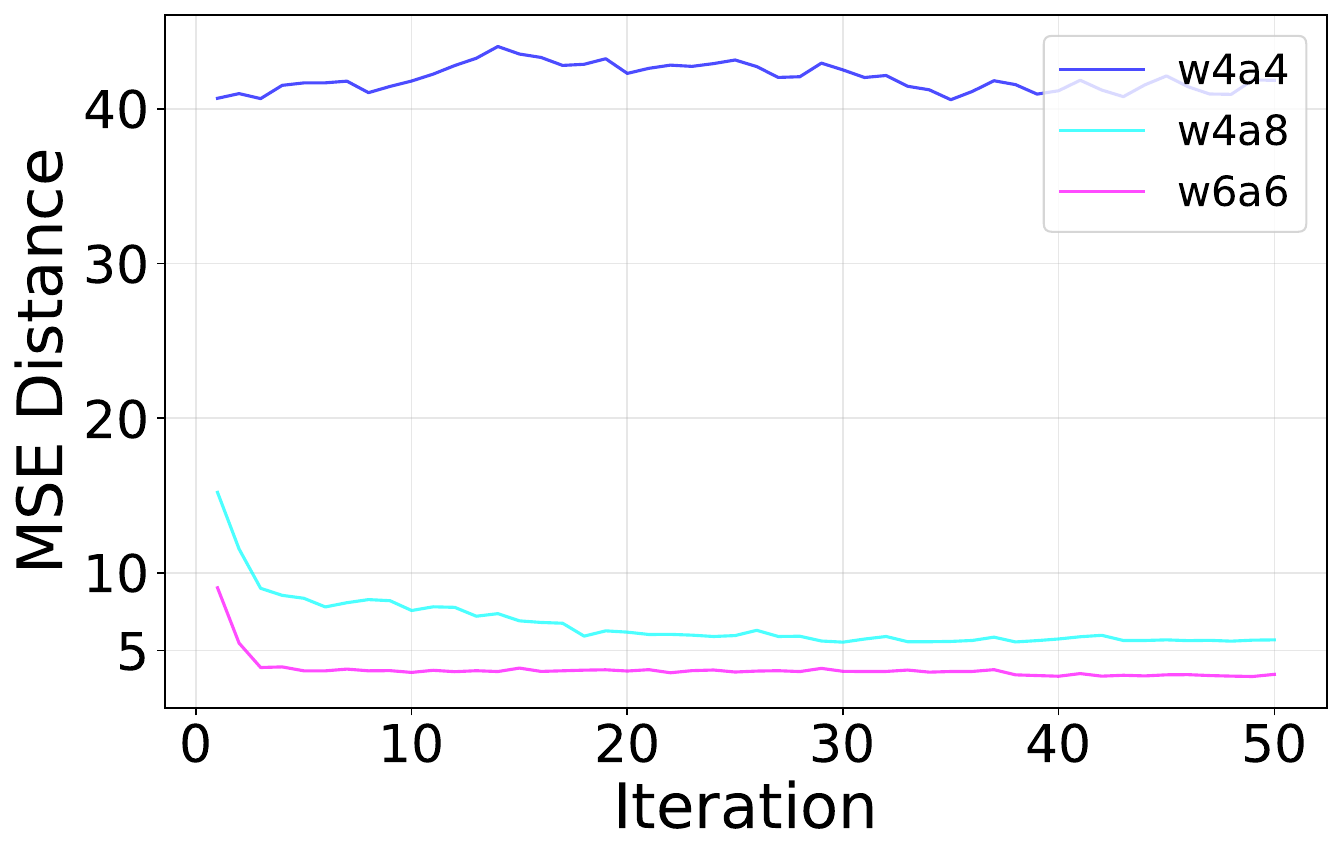}
        \subcaption{SmoothQuant}
    \end{minipage}
    \hspace{-0.01\textwidth}
    \begin{minipage}[b]{0.19\textwidth}
        \includegraphics[width=\textwidth]{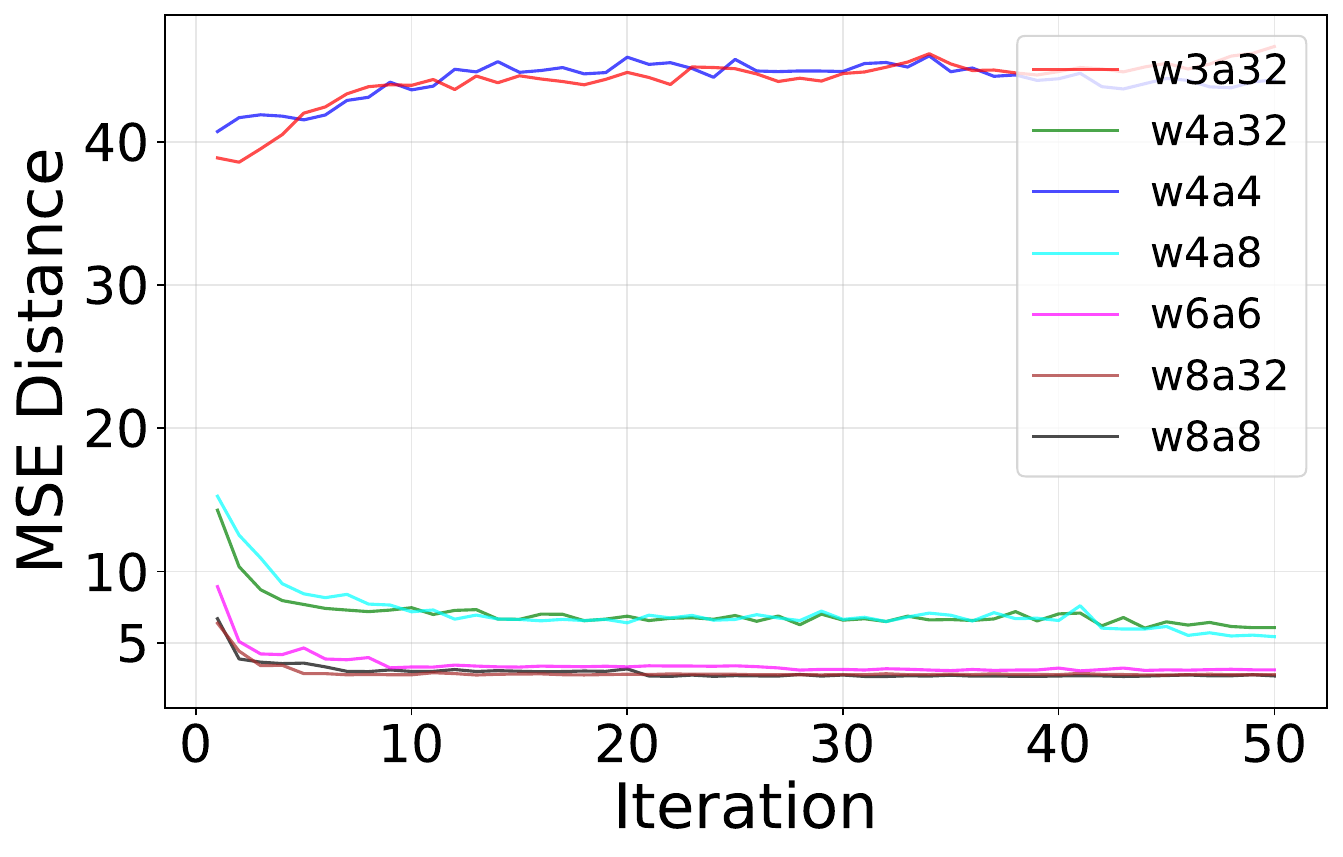}
        \subcaption{OmniQuant}
    \end{minipage}
    \caption{Comparison of the mean squared error (MSE) distance between current and target latent states across varying iterations under different quantization methods and configurations.}
    \label{fig:mse_distance}
\end{figure*}

Figure~\ref{fig:mse_distance} further shows that under aggressive quantization settings, the planning loss often fails to decrease and can even increase over optimization iterations. This behavior suggests a reduced alignment between the planning objective and task success under quantization, making optimization increasingly difficult as precision is reduced.

\section{Key Insights of World Model Quantization}

    \paragraph{Insight~\Romannum{1}:} \textbf{Group-wise Weight Quantization Improves Stability at 4-bit, but Fails under Extreme Precision Reduction.}
    Applying group-wise weight quantization with a moderate group size (e.g., \#G = 128) consistently improves 4-bit performance on Wall and yields moderate gains on PushT, particularly in mid-to-long planning horizons. This suggests that grouping partially alleviates scale mismatch within weight subsets, leading to more stable rollout dynamics.
    However, when weight precision is further reduced (e.g., \#W = 3 on Wall), quantization noise dominates the learned transition dynamics. In this regime, grouping no longer provides meaningful benefits, indicating a failure to preserve useful latent structure under extreme low-bit constraints.
    Group-wise quantization primarily acts as a stabilizer that mitigates scale mismatch in moderately quantized regimes, but it does not fundamentally improve robustness once quantization noise overwhelms the learned dynamics.

    \paragraph{Insight~\Romannum{2}:} \textbf{Per-token Activation Quantization Provides Limited and Inconsistent Gains over Per-tensor.}
    Across both Wall and PushT, per-token activation quantization does not consistently outperform per-tensor quantization in world model planning.
    While more fine-grained scaling can slightly improve performance at moderate bit-widths, these gains are highly sensitive to rollout length and task difficulty.
    In low-bit configurations, per-token quantization can even introduce additional instability, suggesting that increased activation flexibility does not directly translate to improved long-horizon predictions.
    For world models quantization, activation granularity alone is insufficient to guarantee improved planning performance, maintaining stable temporal dynamics appears more critical than maximizing instantaneous representational expressiveness.

    \paragraph{Insight~\Romannum{3}:} \textbf{World Models Exhibit Higher Sensitivity to Representation Quantization.}
    A clear asymmetry emerges between the encoder and predictor modules with respect to quantization sensitivity.
    Quantization applied to the encoder introduces irreversible distortions in the latent representations, which propagate through all subsequent rollout steps and cannot be corrected by additional planning iterations.
    In contrast, predictor quantization primarily affects the temporal consistency of rollouts.
    Its impact is often partially mitigated by longer planning horizons, as the planner can compensate for moderate transition noise through repeated optimization.
    Visual representation is the primary bottleneck in low-bit world models, making encoder precision substantially more critical than predictor precision.

    \paragraph{Insight~\Romannum{4}:} \textbf{PushT Suffers from Planning-level Geometric Misalignment, While Wall Exposes Representation-level Collapse.}
    On PushT, quantization primarily disrupts planning effectiveness rather than visual reconstruction. Across nearly all low-bit settings, success rates drop sharply and cannot be recovered through additional planning iterations. However, qualitative rollouts indicate that reconstructed observations often remain visually plausible as shown in Figure~\ref{fig:rollouts}, suggesting that latent representations are not collapsed. Instead, small but systematic quantization-induced distortions bias the latent dynamics, leading to geometric misalignment between the planned trajectories and the narrow success conditions of the task.
    In contrast, Wall is more sensitive at the representation level. Encoder quantization leads to severe and irreversible degradation in reconstructed observations starting from the initial frame as shown in Figure~\ref{fig:rollouts}. Once corrupted, these representation errors persist throughout the rollout and cannot be corrected by planning, even when quantitative success metrics remain relatively high under moderate quantization.
    Together, these results indicate that low-bit quantization manifests distinct failure modes across tasks, affecting planning effectiveness on PushT and representation fidelity on Wall.

    \paragraph{Insight~\Romannum{5}:} \textbf{Aggressive Low-bit Quantization Disrupts the Planning Objective and Long-horizon Rollout Dynamics.} Under aggressive quantization settings (e.g., Wall with W3 or W4A4, and PushT with W4A8), we consistently observe that the planning loss fails to decrease and even increases over optimization iterations as shown in Figure~\ref{fig:mse_distance}. This behavior suggests that quantization degrades not only predictive accuracy, but also the structure of the rollout dynamics, resulting in a planning objective that is poorly aligned with true task success. Consequently, additional planning iterations yield diminishing or even negative returns. 
    Rather than merely slowing convergence, severe precision constraints induce a mismatch between the learned world model and the planner, rendering optimization-based planning ineffective.


\section{Conclusion}

We presented a systematic empirical study of PTQ for world models in planning-based settings, using DINO-WM as a representative example. Through extensive experiments across multiple quantization methods, quantization granularity, bit-widths, and planning horizons, we showed that quantization effects in world models extend beyond standard accuracy degradation and are tightly coupled with latent rollout dynamics and planning objectives. Our results highlight strong asymmetries in quantization sensitivity across model components, task-dependent failure modes, and the limits of aggressive low-bit precision in long-horizon planning. We hope these findings provide practical guidance for deploying quantized world models under strict computational constraints and motivate future work on quantization strategies that explicitly account for long-horizon planning dynamics.

%
%
\bibliographystyle{splncs04}
\bibliography{main}

@article{zhou2024dino,
  title={Dino-wm: World models on pre-trained visual features enable zero-shot planning},
  author={Zhou, Gaoyue and Pan, Hengkai and LeCun, Yann and Pinto, Lerrel},
  journal={arXiv preprint arXiv:2411.04983},
  year={2024}
}

@article{lecun2022path,
  title={A path towards autonomous machine intelligence version 0.9. 2, 2022-06-27},
  author={LeCun, Yann},
  journal={Open Review},
  volume={62},
  number={1},
  pages={1--62},
  year={2022}
}

@article{zhu2024sora,
  title={Is sora a world simulator? a comprehensive survey on general world models and beyond},
  author={Zhu, Zheng and Wang, Xiaofeng and Zhao, Wangbo and Min, Chen and Li, Bohan and Deng, Nianchen and Dou, Min and Wang, Yuqi and Shi, Botian and Wang, Kai and others},
  journal={arXiv preprint arXiv:2405.03520},
  year={2024}
}

@misc{li2025surveywm,
      title={A Comprehensive Survey on World Models for Embodied AI}, 
      author={Xinqing Li and Xin He and Le Zhang and Min Wu and Xiaoli Li and Yun Liu},
      year={2025},
      eprint={2510.16732},
      archivePrefix={arXiv},
      primaryClass={cs.CV},
      url={https://arxiv.org/abs/2510.16732}, 
}

@misc{ha2018world,
      title={Recurrent World Models Facilitate Policy Evolution}, 
      author={David Ha and Jürgen Schmidhuber},
      year={2018},
      eprint={1809.01999},
      archivePrefix={arXiv},
      primaryClass={cs.LG},
      url={https://arxiv.org/abs/1809.01999}, 
}

@misc{hafner2019planet,
      title={Learning Latent Dynamics for Planning from Pixels}, 
      author={Danijar Hafner and Timothy Lillicrap and Ian Fischer and Ruben Villegas and David Ha and Honglak Lee and James Davidson},
      year={2019},
      eprint={1811.04551},
      archivePrefix={arXiv},
      primaryClass={cs.LG},
      url={https://arxiv.org/abs/1811.04551}, 
}

@misc{hafner2020dreamer,
      title={Dream to Control: Learning Behaviors by Latent Imagination}, 
      author={Danijar Hafner and Timothy Lillicrap and Jimmy Ba and Mohammad Norouzi},
      year={2020},
      eprint={1912.01603},
      archivePrefix={arXiv},
      primaryClass={cs.LG},
      url={https://arxiv.org/abs/1912.01603}, 
}

@misc{vaswani2017attention,
      title={Attention Is All You Need}, 
      author={Ashish Vaswani and Noam Shazeer and Niki Parmar and Jakob Uszkoreit and Llion Jones and Aidan N. Gomez and Lukasz Kaiser and Illia Polosukhin},
      year={2023},
      eprint={1706.03762},
      archivePrefix={arXiv},
      primaryClass={cs.CL},
      url={https://arxiv.org/abs/1706.03762}, 
}

@misc{hafner2023dreamerv3,
      title={Mastering Diverse Domains through World Models}, 
      author={Danijar Hafner and Jurgis Pasukonis and Jimmy Ba and Timothy Lillicrap},
      year={2024},
      eprint={2301.04104},
      archivePrefix={arXiv},
      primaryClass={cs.AI},
      url={https://arxiv.org/abs/2301.04104}, 
}

@misc{ge2024worldgpt,
      title={WorldGPT: Empowering LLM as Multimodal World Model}, 
      author={Zhiqi Ge and Hongzhe Huang and Mingze Zhou and Juncheng Li and Guoming Wang and Siliang Tang and Yueting Zhuang},
      year={2024},
      eprint={2404.18202},
      archivePrefix={arXiv},
      primaryClass={cs.AI},
      url={https://arxiv.org/abs/2404.18202}, 
}

@misc{caron2021dino,
      title={Emerging Properties in Self-Supervised Vision Transformers}, 
      author={Mathilde Caron and Hugo Touvron and Ishan Misra and Hervé Jégou and Julien Mairal and Piotr Bojanowski and Armand Joulin},
      year={2021},
      eprint={2104.14294},
      archivePrefix={arXiv},
      primaryClass={cs.CV},
      url={https://arxiv.org/abs/2104.14294}, 
}

@misc{oquab2024dinov2,
      title={DINOv2: Learning Robust Visual Features without Supervision}, 
      author={Maxime Oquab and Timothée Darcet and Théo Moutakanni and Huy Vo and Marc Szafraniec and Vasil Khalidov and Pierre Fernandez and Daniel Haziza and Francisco Massa and Alaaeldin El-Nouby and Mahmoud Assran and Nicolas Ballas and Wojciech Galuba and Russell Howes and Po-Yao Huang and Shang-Wen Li and Ishan Misra and Michael Rabbat and Vasu Sharma and Gabriel Synnaeve and Hu Xu and Hervé Jegou and Julien Mairal and Patrick Labatut and Armand Joulin and Piotr Bojanowski},
      year={2024},
      eprint={2304.07193},
      archivePrefix={arXiv},
      primaryClass={cs.CV},
      url={https://arxiv.org/abs/2304.07193}, 
}

@article{liu2021post,
  title={Post-training quantization for vision transformer},
  author={Liu, Zhenhua and Wang, Yunhe and Han, Kai and Zhang, Wei and Ma, Siwei and Gao, Wen},
  journal={Advances in Neural Information Processing Systems},
  volume={34},
  pages={28092--28103},
  year={2021}
}

@article{fu2020d4rl,
  title={D4rl: Datasets for deep data-driven reinforcement learning},
  author={Fu, Justin and Kumar, Aviral and Nachum, Ofir and Tucker, George and Levine, Sergey},
  journal={arXiv preprint arXiv:2004.07219},
  year={2020}
}

@article{tassa2018deepmind,
  title={Deepmind control suite},
  author={Tassa, Yuval and Doron, Yotam and Muldal, Alistair and Erez, Tom and Li, Yazhe and Casas, Diego de Las and Budden, David and Abdolmaleki, Abbas and Merel, Josh and Lefrancq, Andrew and others},
  journal={arXiv preprint arXiv:1801.00690},
  year={2018}
}

@article{brooks2024video,
  title={Video generation models as world simulators},
  author={Brooks, Tim and Peebles, Bill and Holmes, Connor and DePue, Will and Guo, Yufei and Jing, Li and Schnurr, David and Taylor, Joe and Luhman, Troy and Luhman, Eric and others},
  journal={OpenAI Blog},
  volume={1},
  number={8},
  pages={1},
  year={2024}
}

@inproceedings{wu2023daydreamer,
  title={Daydreamer: World models for physical robot learning},
  author={Wu, Philipp and Escontrela, Alejandro and Hafner, Danijar and Abbeel, Pieter and Goldberg, Ken},
  booktitle={Conference on robot learning},
  pages={2226--2240},
  year={2023},
  organization={PMLR}
}

@misc{gupta2015deeplearninglimitednumerical,
      title={Deep Learning with Limited Numerical Precision}, 
      author={Suyog Gupta and Ankur Agrawal and Kailash Gopalakrishnan and Pritish Narayanan},
      year={2015},
      eprint={1502.02551},
      archivePrefix={arXiv},
      primaryClass={cs.LG},
      url={https://arxiv.org/abs/1502.02551}, 
}

@misc{nagel2021whitepaper,
      title={A White Paper on Neural Network Quantization}, 
      author={Markus Nagel and Marios Fournarakis and Rana Ali Amjad and Yelysei Bondarenko and Mart van Baalen and Tijmen Blankevoort},
      year={2021},
      eprint={2106.08295},
      archivePrefix={arXiv},
      primaryClass={cs.LG},
      url={https://arxiv.org/abs/2106.08295}, 
}

@misc{gholami2021quantsurvey,
      title={A Survey of Quantization Methods for Efficient Neural Network Inference}, 
      author={Amir Gholami and Sehoon Kim and Zhen Dong and Zhewei Yao and Michael W. Mahoney and Kurt Keutzer},
      year={2021},
      eprint={2103.13630},
      archivePrefix={arXiv},
      primaryClass={cs.CV},
      url={https://arxiv.org/abs/2103.13630}, 
}

@article{krishnamoorthi2018rtn,
  title={Quantizing deep convolutional networks for efficient inference: A whitepaper}, 
  author={Raghuraman Krishnamoorthi},
  journal={arXiv preprint arXiv:1806.08342},
  year={2018},
}

@misc{li2021brecq,
      title={BRECQ: Pushing the Limit of Post-Training Quantization by Block Reconstruction}, 
      author={Yuhang Li and Ruihao Gong and Xu Tan and Yang Yang and Peng Hu and Qi Zhang and Fengwei Yu and Wei Wang and Shi Gu},
      year={2021},
      eprint={2102.05426},
      archivePrefix={arXiv},
      primaryClass={cs.LG},
      url={https://arxiv.org/abs/2102.05426}, 
}

@misc{wei2023qdrop,
      title={QDrop: Randomly Dropping Quantization for Extremely Low-bit Post-Training Quantization}, 
      author={Xiuying Wei and Ruihao Gong and Yuhang Li and Xianglong Liu and Fengwei Yu},
      year={2023},
      eprint={2203.05740},
      archivePrefix={arXiv},
      primaryClass={cs.CV},
      url={https://arxiv.org/abs/2203.05740}, 
}

@misc{yuan2024ptq4vit,
      title={PTQ4ViT: Post-training quantization for vision transformers with twin uniform quantization}, 
      author={Zhihang Yuan and Chenhao Xue and Yiqi Chen and Qiang Wu and Guangyu Sun},
      year={2024},
      eprint={2111.12293},
      archivePrefix={arXiv},
      primaryClass={cs.CV},
      url={https://arxiv.org/abs/2111.12293}, 
}

@misc{liu2023noisyquant,
      title={NoisyQuant: Noisy Bias-Enhanced Post-Training Activation Quantization for Vision Transformers}, 
      author={Yijiang Liu and Huanrui Yang and Zhen Dong and Kurt Keutzer and Li Du and Shanghang Zhang},
      year={2023},
      eprint={2211.16056},
      archivePrefix={arXiv},
      primaryClass={cs.CV},
      url={https://arxiv.org/abs/2211.16056}, 
}

@misc{lin2024duquant,
      title={DuQuant: Distributing Outliers via Dual Transformation Makes Stronger Quantized LLMs}, 
      author={Haokun Lin and Haobo Xu and Yichen Wu and Jingzhi Cui and Yingtao Zhang and Linzhan Mou and Linqi Song and Zhenan Sun and Ying Wei},
      year={2024},
      eprint={2406.01721},
      archivePrefix={arXiv},
      primaryClass={cs.CL},
      url={https://arxiv.org/abs/2406.01721}, 
}

@misc{sun2025flatquant,
      title={FlatQuant: Flatness Matters for LLM Quantization}, 
      author={Yuxuan Sun and Ruikang Liu and Haoli Bai and Han Bao and Kang Zhao and Yuening Li and Jiaxin Hu and Xianzhi Yu and Lu Hou and Chun Yuan and Xin Jiang and Wulong Liu and Jun Yao},
      year={2025},
      eprint={2410.09426},
      archivePrefix={arXiv},
      primaryClass={cs.CL},
      url={https://arxiv.org/abs/2410.09426}, 
}

@inproceedings{choukroun2019low,
  title={Low-bit quantization of neural networks for efficient inference},
  author={Choukroun, Yoni and Kravchik, Eli and Yang, Fan and Kisilev, Pavel},
  booktitle={2019 IEEE/CVF International Conference on Computer Vision Workshop (ICCVW)},
  pages={3009--3018},
  year={2019},
  organization={IEEE}
}

@article{lin2024awq,
  title={Awq: Activation-aware weight quantization for on-device llm compression and acceleration},
  author={Lin, Ji and Tang, Jiaming and Tang, Haotian and Yang, Shang and Chen, Wei-Ming and Wang, Wei-Chen and Xiao, Guangxuan and Dang, Xingyu and Gan, Chuang and Han, Song},
  journal={Proceedings of machine learning and systems},
  volume={6},
  pages={87--100},
  year={2024}
}

@inproceedings{xiao2023smoothquant,
  title={Smoothquant: Accurate and efficient post-training quantization for large language models},
  author={Xiao, Guangxuan and Lin, Ji and Seznec, Mickael and Wu, Hao and Demouth, Julien and Han, Song},
  booktitle={International conference on machine learning},
  pages={38087--38099},
  year={2023},
  organization={PMLR}
}

@article{shao2023omniquant,
  title={Omniquant: Omnidirectionally calibrated quantization for large language models},
  author={Shao, Wenqi and Chen, Mengzhao and Zhang, Zhaoyang and Xu, Peng and Zhao, Lirui and Li, Zhiqian and Zhang, Kaipeng and Gao, Peng and Qiao, Yu and Luo, Ping},
  journal={arXiv preprint arXiv:2308.13137},
  year={2023}
}
\end{document}